\documentclass[twocolumn,journal]{IEEEtran}
\usepackage{amsfonts}
\usepackage{amssymb} 
\usepackage{dsfont}
\usepackage{lipsum}
\usepackage{graphicx}
\usepackage{amsmath}
\usepackage{graphics}
\usepackage{cite}    
\usepackage{bigstrut}
\usepackage{xcolor,colortbl}
\usepackage{float}
\usepackage{balance}
\usepackage{verbatim}
\usepackage{lipsum}
\usepackage{psfrag}
\usepackage{multirow}
\usepackage{bbm}
\usepackage{multicol}

\usepackage{enumitem}

\usepackage{alphabeta}
\usepackage{amsmath,amssymb}
\usepackage{verbatim}
\usepackage{float}
\usepackage{steinmetz}
\usepackage{hyperref}
\usepackage{caption}
\usepackage{amssymb}  
\usepackage{pifont}   
\usepackage{float}
\usepackage{mathtools}
\usepackage{algpseudocode}
\usepackage{algorithm}
\usepackage{mathtools, nccmath}
\usepackage{diagbox}
\usepackage{subcaption}
\usepackage{stackengine}
\definecolor{lightgray}{rgb}{0.75,0.75 ,0.75 }
\newcolumntype{P}[1]{>{\centering\arraybackslash}p{#1}}

\usepackage{afterpage}
\usepackage{tikz}
\usepackage{mathdots}
\usepackage{yhmath}
\usepackage{cancel}
\usepackage{color}
\usepackage{siunitx}
\usepackage{array}
\usepackage{multirow}
\usepackage{textcomp}
\usepackage{gensymb}
\usepackage{tabularx}
\usepackage{booktabs}
\usepackage{graphicx}
\usepackage{xcolor}
\usepackage{pgfplots}
\usetikzlibrary{external}

\begin{document}
	\IEEEoverridecommandlockouts
	
\title{Heterogeneous Resource Allocation with Multi-task Learning for Wireless Networks}
\author{Nikos A. Mitsiou,~\IEEEmembership{Graduate Student Member,~IEEE}, 
        Pavlos S. Bouzinis, \\
        Panagiotis G. Sarigiannidis,~\IEEEmembership{Senior Member,~IEEE}, 
        and George K. Karagiannidis,~\IEEEmembership{Fellow,~IEEE}%
\thanks{Nikos A. Mitsiou, and George K. Karagiannidis are with the Wireless Communications and Information Processing (WCIP) Group, Department of Electrical and Computer Engineering, Aristotle University of Thessaloniki, 54124 Thessaloniki, Greece (e-mails: nmitsiou@auth.gr, geokarag@auth.gr).}
\thanks{Pavlos S. Bouzinis is with MetaMind Innovations P.C., 50100 Kozani, Greece (e-mail: pbouzinis@metamind.gr).}
\thanks{Panagiotis G. Sarigiannidis is with the Department of Informatics and Telecommunications Engineering, University of Western Macedonia, 50100 Kozani, Greece (e-mail: psarigiannidis@uowm.gr).}
\thanks{This work has received funding from the Smart Networks and Services Joint Undertaking (SNS JU) under the European Union's Horizon Europe research and innovation programme (Grant Agreement No. 101096456 - NANCY).}
\thanks{We note that part of this work was presented in the IEEE International Symposium on Personal, Indoor and Mobile Radio Communications (PIMRC), Valencia, Spain, 2024 \cite{mitsiou2024multi}.}
}
\maketitle
\begin{abstract}
\textcolor{black}{The optimal solution to an optimization problem depends on the problem's objective function, constraints, and size. While deep neural networks (DNNs) have proven effective in solving optimization problems, changes in the problem's size, objectives, or constraints often require adjustments to the DNN architecture to maintain effectiveness, or even retraining a new DNN from scratch. Given the dynamic nature of wireless networks, which involve multiple and diverse objectives that can have conflicting requirements and constraints, we propose a multi-task learning (MTL) framework to enable a single DNN to jointly solve a range of diverse optimization problems. In this framework, optimization problems with varying dimensionality values, objectives, and constraints are treated as distinct tasks. To jointly address these tasks, we propose a conditional computation-based MTL approach with routing. The multi-task DNN consists of two components, the base DNN (bDNN), which is the single DNN used to extract the solutions for all considered optimization problems, and the routing DNN (rDNN), which manages which nodes and layers of the bDNN to be used during the forward propagation of each task. The output of the rDNN is a binary vector which is multiplied with all bDNN's weights during the forward propagation, creating a unique computational path through the bDNN for each task. This setup allows the tasks to either share parameters or use independent ones, with the decision controlled by the rDNN. The proposed framework supports both supervised and unsupervised learning scenarios. Numerical results demonstrate the efficiency of the proposed MTL approach in solving diverse optimization problems. In contrast, benchmark DNNs lacking the rDNN mechanism were unable to achieve similar levels of performance, highlighting the effectiveness of the proposed architecture.}

\end{abstract}
\vspace{-0cm}
\maketitle
\section{Introduction}
Deep learning (DL) has become a key facilitator in integrating intelligent features into the physical (PHY) and medium-access control (MAC) layers of wireless networks \cite{marco, proc}. By utilizing deep neural networks (DNNs) architectures and large datasets, DL empowers wireless networks to dynamically adjust and enhance their performance based on the considered key performance indicators (KPIs) \cite{marco}. Furthermore, model-free techniques are anticipated to entirely, or to some extent, replace traditional optimization methods.  This is because model-free approaches can better adapt to complex and high-dimensional systems where explicit modeling is challenging \cite{dimakis}. This is essentially true in wireless communications, where acquiring accurate models that can be used to predict the network dynamics and their impact on performance is often infeasible \cite{zorzi}.
\textcolor{black}{DL-based methods  have the advantage of delivering nearly real-time decisions.
Once trained, only a forward pass is needed whenever the
network’s parameters change. In contrast, model-based methods require a rerun from scratch, while certain conditions,
such as convexity, need to hold. Thus, despite the potentially higher theoretical complexity of DL-based methods, their practical benefits in terms of scalability and adaptability to non-convex and heterogeneous tasks, make them advantageous compared to  numerical methods \cite{marco}.}

However, both model-based and the conventional DL-based methods encounter limitations as they can usually tackle optimization problems of fixed dimensionality and objectives, while both approaches need to be reinitiated, if the dimensionality or the objective of the optimization problem changes \cite{akrout2023multilayer}.  Multi-task learning (MTL) is a promising machine learning (ML) paradigm to address this issue, which involves training models with data from multiple tasks simultaneously, using shared representations to capture common ideas among related tasks. MTL methods are often categorized into two groups: hard parameter sharing and soft parameter sharing. In hard parameter sharing, model weights are shared between tasks, aiming to jointly minimize multiple loss functions. In soft parameter sharing, tasks have individual models with separate weights, but the joint objective function includes the distance between the trainable parameters of different tasks. MTL has found success in various tasks, including image processing, robotics manipulation, and etc. \cite{mt, routing, modular, once, dynamic, slimmable}. Despite its potential, the research community has yet to explore the application of MTL for jointly addressing optimization problems with varying dimensionality and objectives. Consequently, the feasibility of leveraging MTL to enhance wireless communication networks remains uncertain.
\subsection{Related Work}
\textcolor{black}{ DNNs have  been widely used for tackling various complex optimization problems, especially in the context of wireless communications \cite{d2,d4,d8,online2,online1}.  In \cite{d2,d4,d8} distributed and unsupervised learning (UL) based frameworks for constrained optimization were proposed. In \cite{online1}, online-learning was used to tackle optimization problems in intelligent reflecting surface (IRS) networks, achieving greater performance against model-based methods. Moreover, meta-learning has emerged as a promising ML framework which enables DNNs to quickly adapt to new unseen tasks  using limited data \cite{rev4, rev6}.  Despite their novelty, the works of \cite{rev4, rev6}  cannot been applied to problems of varying dimensionality, while they also require that  the considered optimization tasks have the same set of constraints, which is in contrast to the MTL scheme proposed in this paper.}
Another  technique which aims to enhance the ability of DNNs to generalize their performance to multiple tasks is the zero-padding (ZP).
For instance, in \cite{z1, z10, z12}, ZP was used to handle the changing dimensionality of the input data due to the dynamic nature of wireless networks. 

Furthermore, a promising research direction which has shown great capability to address the scalability issues of wireless resource allocation is graph neural networks (GNNs) \cite{gnn5}. In one approach called the random edge GNN (REGNN) \cite{gnn1}, convolutions were applied over random graphs representing fading interference patterns in wireless networks. Additionally, GNNs have been applied to solve problems like optimal power control, beamforming, and user selection \cite{gnn2,gnn3}. These studies modelled these tasks using graphs and solve them effectively with GNNs. Moreover,  
GNNs have been employed along with primal-dual optimization techniques \cite{gnn4} to perform optimal constrained resource allocation. Furthermore, GNNs have been used to learn power allocation in multi-cell-multi-user systems with heterogeneous GNNs \cite{gnn6}.  Nonetheless, there are only a few studies that have explored MTL with GNNs  \cite{mtgn1,mtgn2}, \textcolor{black}{ but not in the context of multi-task constrained optimization.} These approaches rely on techniques such as soft parameter sharing or regularization between task-specific networks, which increase the total number of trainable parameters compared to a single-task DNN \cite{mtgn1,mtgn2}. 

Moreover,  evolutionary multi-tasking optimization (EMTO) aims at solving multiple optimization problems simultaneously \cite{MTO1}.
In EMTO, a population of candidate solutions evolves over generations, with each individual tasked with solving a specific optimization problem. 
Through the process of natural selection and mutation the population evolves to produce solutions that are adept at solving all the given tasks \cite{MTO2}. Therefore, EMTO algorithms utilize the underlying similarity of optimization tasks to efficiently solve heterogeneous tasks just like MTL \cite{MTO3}. Moreover, in \cite{MTO4,MTO5}, EMTO algorithms for constrained multi-task optimization problems were proposed which helped generalize EMTO to practical use cases. Nonetheless, EMTO is time-consuming and cannot be used to solve real-time problems in dynamic environments, such as wireless networks, where the parameters of the optimization problems change constantly.

\subsection{Motivation \& Contribution}
\textcolor{black}{
Despite their novelty, existing studies on resource allocation have not thoroughly examined a comprehensive framework capable of addressing multiple optimization problems simultaneously, each with different objectives, constraints and dimensionality. For instance, GNNs have been shown to be a scalable solution which can solve problems of different dimensionality, but they cannot be applied to problems of different objective or constraints. In addition, meta-learning can help to learn new unseen tasks but it requires that the tasks which it is trained on have the same set of constraints and dimensionality, \textcolor{black}{while ZP cannot manage the interference that arises during the joint training of multiple, different tasks.} We note that such a framework can be particularly beneficial, in future wireless networks, where various key performance indicators (KPIs), each one described by a distinct objective function and constraints, will need to be simultaneously considered to meet the diverse and often contradictory, requirements of all users.} As such, in the authors' perspective, introducing such a framework not only aligns with the evolution of wireless resource allocation literature, but it has practical motivation too.  For instance, in the context of Open-RAN, a MTL solution can allow for the consolidation of multiple resource allocation tasks within a single xApp, simplifying the orchestration of network functions and reducing overhead associated with the life-cycle management of xApps \cite{oran}.

As such, we tackle this challenge by adopting a MTL approach. Our primary focus is to determine the effectiveness of MTL in jointly solving problems of diverse objectives, constraints and varying dimensionality within wireless networks. \textcolor{black}{To this end, we treat optimization problems of different objective functions and constraints, as distinct tasks. Also, we consider optimization problems with the same objective but different dimensionality, as separate tasks.} Subsequently, we propose a dynamic DNN, based on MTL, which is capable of effectively capturing this mapping, and it is applicable to both SL and UL. To achieve this, we employ conditional computation with routing, which involves two DNN components, the base DNN (bDNN) and the router DNN (rDNN). The bDNN is the single DNN used for the forward propagation of all tasks, while the rDNN guides each task through a different computational path of the bDNN and enables the bDNN to generalize its performance to all tasks. The contribution of the paper is summarized below:
\begin{itemize}
    \item A multi-task mapping is defined to address optimization problems of varying dimensionalities and diverse objectives. A DNN architecture based on MTL is then proposed to efficiently capture this mapping. \textcolor{black}{Notably, MTL for optimization problems of both different dimensionality and varying objective functions or constraint sets has not been thoroughly evaluated in the literature.}
    \item To realize the proposed multi-task DNN, the concept of conditional computation with routing is implemented. This involves the utilization of the bDNN, responsible for the inference procedure of all tasks, and the rDNN, which selects the computational path through the bDNN for each distinct task. 
    \textcolor{black}{The design of the rDNN is based on \emph{hard parameter sharing}, thus its output is binary and it is multiplied with the bDNN's weights, to jointly optimize and configure the paths of all tasks. In addition, hard parameter sharing does not increase the number of parameters of the bDNN, enabling a fair comparison between the proposed scheme and any single-task DNN.}
    \item The proposed multi-task DNN is evaluated under fourteen different allocation tasks. The numerical results demonstrate that its performance is near the performance of the optimal single-task DNN, which is trained for each optimization task independently. In contrast, benchmark DNNs lacking an rDNN failed to achieve similar success further showcasing the effectiveness of the proposed rDNN architecture. 
\end{itemize}

\section{The Single-task Problem Formulation}
First, we consider the set $\mathcal{N} = \{1,2,...,N\}$, and the following optimization problem 
\begin{equation} \tag{$\textbf{P}_1$}\label{eq:original}
\begin{aligned}
\underset{{\boldsymbol{x}}}{\text{min}} \quad &f_0\left(\boldsymbol{x};\boldsymbol{a}\right) \\
\quad \text{s.t.}\quad   &f_n\left(\boldsymbol{x};\boldsymbol{a}\right)\leq 0, \,\,\, \forall n \in \mathcal{N}, \\
\quad&\boldsymbol{x}\in \mathcal{X},
\end{aligned}
\end{equation}
where the function $f_0:\mathbb{R}^N \rightarrow \mathbb{R}$ describes the network's cost function and the functions $f_1, ..., f_N:\mathbb{R}^N \rightarrow \mathbb{R}$ indicate the devices local constraints. The functions $f_0,f_1,..,f_N$ are not necessarily convex in the general case, but are differentiable. The set $\mathcal{X} \subseteq \mathbb{R}^N$ is a nonempty, compact, and convex set, reflecting the set of global constraints.  Let us denote the optimal solution of problem \eqref{eq:original} as $\boldsymbol{x}^{*}(\boldsymbol{a})\in\mathcal{X}$. The optimal solution is parameterized by the parameter $\boldsymbol{a}\in\mathcal{A}\subseteq \mathbb{R}^N$. 
Thus, there exists an optimal mapping between the parameter set $\mathcal{A}$ and the set $\mathcal{X}^{*}$ which is the set containing all optimal solutions of problem \eqref{eq:original}, i.e.,
\begin{equation}
\begin{aligned}\label{eq:opt_set}
    \mathcal{X}^{*} \triangleq &\Big\{\boldsymbol{x}^{*}\in\mathcal{X}\,\Big| \,f_0\left(\boldsymbol{x}^{*};\boldsymbol{a}\right) \leq f_0\left(\boldsymbol{x};\boldsymbol{a}\right), \forall \boldsymbol{x}\in\mathcal{X},\\
    &f_n\left(\boldsymbol{x}^*;\boldsymbol{a}\right)\leq 0, \, \forall n \in \mathcal{N}\Big\}.
\end{aligned}
\end{equation}
This mapping will be denoted as $\mathcal{F}$ and is given below
\begin{equation} \label{mapping}
\mathcal{F}: \,\, \mathcal{A} \xrightarrow[]{\text{\ref{eq:original}}} \mathcal{X}^{*}.
\end{equation}
Essentially, finding the mapping $\mathcal{F}$ that provides the optimal set of solutions for problem \eqref{eq:original}, can be conceived as an individual \textit{task} with input and output dimensionality of $\mathbb{R}^N$, where we define the input of the task to be the parameters $\boldsymbol{a}$ and its output to be the optimal value $\boldsymbol{x}^{*}(\boldsymbol{a})$. \footnote{It is clarified that the paper can be generalized to the case that the problem \eqref{eq:original} has different input and output dimensionality, or to the case where both continuous and discrete variables exist. } 
Next, we present two conventional DL approaches for approximately obtaining the desired mapping $\mathcal{F}$.
\subsection{Supervised Learning} \label{section:supervised}
We assume that $f_1, ..., f_N$ are convex functions. Then, problem \eqref{eq:original} can be solved optimally for any $\boldsymbol{a}$ by utilizing convex optimization tools, and the optimal solution $\boldsymbol{x}^{*}(\boldsymbol{a})$ is obtained. Nonetheless, problem \eqref{eq:original} should be resolved whenever input parameter $\boldsymbol{a}$ undergoes a change, while this process may be time-consuming.  In wireless communication networks, the input parameter $\boldsymbol{a}$ may change frequently, thus model-based methods cannot satisfy real-time constraints \cite{dimakis}. SL has demonstrated its efficacy in dealing with resource allocation problems, while once trained, it requires only a forward pass, rendering it suitable for real-time resource management \cite{8642915}.

Let us assume a feed-forward fully-connected DNN, with $L+1$ layers and $d_l$ neurons per layer. Let $\boldsymbol{y}^{l-1}$ to be the input to the $l$-th layer of the network, with $\boldsymbol{y}^0$ being the input to the input layer of the DNN, while in our case it holds that $\boldsymbol{y}^0=\boldsymbol{a}$. Then, for all $l=1,...,L+1$ and $n=1,...,d_l$ the output $\boldsymbol{y}^l $ of node $n$ in layer $l$ is obtained as
\begin{equation}
    \boldsymbol{y}^l  = g_{n,l}(z_{n,l}),\,\,z_{n,l} = \boldsymbol{w}_{n,l}^\intercal\boldsymbol{y}^{l-1}+b_{n,l}
\end{equation}
wherein $\boldsymbol{w}_{n,l}\in \mathbb{R}^{d_{l-1}}$ with $\boldsymbol{w}_{n,l}(k)$ being the weight of the
link between the $k$-th neuron in layer $l-1$ and the $n$-th neuron in layer $l$, $b_{n,l}\in \mathbb{R}$ is the bias term of neuron $n$ in layer $l$, while $g_{n,l}$, is the activation function of neuron $n$ in layer $l$. 
Also, $\boldsymbol{\Theta}\triangleq\{\boldsymbol{W}_l,\boldsymbol{b}_l\}_{l=1}^{L+1}$ denotes the weights and biases of all layers, i.e., the training parameters, $\boldsymbol{W}_l\in\mathbb{R}^{d_{l-1} \times d_l}$ is the matrix containing all the weights between the $(l-1)$-th and the $l$-th hidden layer, and $\boldsymbol{b}_l\in\mathbb{R}^{d_l}$ is the vector which contains all the biases of the $l$-th layer.  

Since problem \eqref{eq:original} can be optimally solved $\forall \boldsymbol{a}\in\mathcal{A}$, a dataset $\mathcal{D}=\{\boldsymbol{a}^u,\boldsymbol{x}^{*}(\boldsymbol{a}^u)\}_{u=1}^{D}$, which consists of  $D=\lvert\mathcal{D}\rvert$ samples, with data input $\boldsymbol{a}^u$ along with their respected label/target $\boldsymbol{x}^{*}(\boldsymbol{a}^u)$, is generated. \textcolor{black}{To evaluate the proposed framework we consider the dataset to be perfect without missing or corrupted data. The robustness of the proposed framework under missing or corrupted data is left as a future work.} Therefore, the aim of the SL approach is to obtain a $\boldsymbol{y}^{L+1}\left(\boldsymbol{\Theta};\boldsymbol{a}\right) \approx \boldsymbol{x}^{*}(\boldsymbol{a}),\forall \boldsymbol{a}\in\mathcal{A}$, where $\boldsymbol{y}^{L+1}\left(\boldsymbol{\Theta};\boldsymbol{a}\right)$ is the output of the final layer of the DNN. Then, the SL approach is to train a DNN to undertake the task described from the mapping $\mathcal{F}$ of \eqref{mapping}, with respect to the dataset $\mathcal{D}$. To this end, the loss function of the SL method is defined as 
\begin{equation} \label{loss_super}
    L(\boldsymbol{\Theta}) = \frac{1}{B}\sum_{u \in \mathcal {B}} \mathcal{L}\left(\boldsymbol{x}^{*}(\boldsymbol{a}^u),\boldsymbol{y}^{L+1}\left(\boldsymbol{\Theta};\boldsymbol{a}^u\right)\right),
\end{equation}
where $\mathcal{B}\subseteq\mathcal{D}$ and $B=\lvert\mathcal{B}\rvert$ is the batch size of the training dataset, while the loss function depends on both the dataset's inputs and targets. 
A common choice for the loss function, which captures the similarity between the target $\boldsymbol{x}^{*}(\boldsymbol{a}^u)$ for the $i$-th sample, and the output $\boldsymbol{y}^{L+1}(\boldsymbol{\Theta};\boldsymbol{a}^u)$ of the DNN is the minimum square error (MSE) function, which is given below
\begin{equation}
    \mathcal{L}\left(\cdot,\cdot\right) = \left(\boldsymbol{x}^{*}(\boldsymbol{a}^u)-\boldsymbol{y}^{L+1}\left(\boldsymbol{\Theta};\boldsymbol{a}^u\right)\right)^2.
\end{equation}
However, in order to maintain generality, the loss function in \eqref{loss_super} will be used in the remainder of the paper. 
\subsection{Unsupervised Learning}
Model-based methods for solving problem  \eqref{eq:original}, and subsequently, SL, require that some conditions for $f_0,...,f_N$, such as convexity, hold. This is mandatory for obtaining $\boldsymbol{x}^{*}(\boldsymbol{a})$, $\forall \boldsymbol{a}\in\mathcal{A}$. However, several fundamental problems in the field of wireless communications are non-convex. These problems often are not solvable by standard optimization tools \cite{d4}. For this reason, UL is a promising approach to overcome this challenge \cite{d4,marco} and solve problems of the form given in \eqref{eq:original}, when conditions such as convexity do not hold. 
We note that, in \cite{d4,amos2017optnet}, it was shown that any convex projection on $\mathcal{X}$
can be realized via DNNs. For instance, an often-encountered constraint is of the form $\boldsymbol{1}^\intercal\boldsymbol{x} \leq 1$, which can be handled by using the Softmax function as the output layer of the DNN. Nonetheless, in the case that the projection onto set $\mathcal{X}$ is not convex, or it is not straightforward to be implemented via the DNN, we can simply concatenate the constraints imposed by the set $\mathcal{X}$ into the rest inequality constraints of \eqref{eq:original} and proceed without needing to project $\boldsymbol{x}$ onto the set $\mathcal{X}$ explicitly. Thus, hereinafter, without loss of generality, the constraint $x\in\mathcal{X}$ of problem \eqref{eq:original} will be discarded.  Following subsection II.\ref{section:supervised}, we consider a feedforward DNN, thus the output of the unsupervised DNN is given as $\boldsymbol{y}^{L+1}\left(\boldsymbol{\Theta};\boldsymbol{a}\right)$. 

The aim of the  UL approach is to obtain a $\boldsymbol{y}^{L+1}\left(\boldsymbol{\Theta};\boldsymbol{a}\right)$, so that $\boldsymbol{y}^{L+1}\left(\boldsymbol{\Theta};\boldsymbol{a}\right) \approx \boldsymbol{x}^{*}(\boldsymbol{a})$, $\forall \boldsymbol{a}\in\mathcal{A}$. However, in contrast to the SL case, the dataset $\mathcal{D}=\{\boldsymbol{a}^u,\boldsymbol{x}^{*}(\boldsymbol{a}^u)\}_{u=1}^{D}$ cannot be created since there is no tractable way to obtain the values $\boldsymbol{x}^{*}(\boldsymbol{a}^u)$ for each $\boldsymbol{a}^u$. \textcolor{black}{We note that numerical methods could possibly be employed to construct a dataset similar to the SL case, however, the results from [13] demonstrate the superior performance of UL-based optimization over numerical methods, particularly in non-convex settings.} Therefore, in the UL case, the dataset will be given as $\mathcal{D}=\{\boldsymbol{a}^u\}_{u=1}^{D}$, and the UL-based DNN has to be trained in such a fashion so that $\boldsymbol{y}^{L+1}\left(\boldsymbol{\Theta};\boldsymbol{a}^u\right) \approx \boldsymbol{x}^{*}(\boldsymbol{a}^u)$, $\forall \boldsymbol{a}^u\in\mathcal{D}$.
To this end, it is imperative to define the loss function of the UL-based DNN so that it includes both the objective function of problem \eqref{eq:original}, and its constraints. A possible approach to address this is the Lagrange duality method \cite{d4}. First, the Lagrangian of \eqref{eq:original} is given as follows
\begin{equation}
    \mathcal{L}_g(\boldsymbol{x},\boldsymbol{\lambda}) = f_0\left(\boldsymbol{x};\boldsymbol{a}\right) + \sum_{n\in\mathcal{N}} \lambda_nf_n\left(\boldsymbol{\boldsymbol{x};\boldsymbol{a}}\right),
\end{equation}
where the non-negative $\lambda_n$ corresponds to the dual variable associated with the $n$-th constraint in \eqref{eq:original}. $\boldsymbol{\lambda}\in\mathbb{R}^N$ denotes the vector containing all $\lambda_n, \forall n$. Moreover, the dual function $\mathcal{G}(\boldsymbol{\lambda})$ is given as
\begin{equation}
    \mathcal{G}(\boldsymbol{\lambda}) = \underset{\boldsymbol{x}}{\mathrm{inf}}\,\mathcal{L}_g(\boldsymbol{x},\boldsymbol{\lambda}),
\end{equation}
while the respected dual problem is defined as
\begin{equation} \tag{$\textbf{P}_{2}$}\label{eq:problem2.1l}
\begin{aligned}
\underset{{\boldsymbol{\lambda}}}{\text{max}} \quad &\mathcal{G}(\boldsymbol{\lambda}) \\
\quad \text{s.t.}\quad   &\boldsymbol{\lambda}\succeq 0
\end{aligned}
\end{equation} 
To solve problem \eqref{eq:problem2.1l} the primal-dual method can be employed \cite{d4}. By taking into account that $\boldsymbol{x} = \boldsymbol{y}^{L+1}\left(\boldsymbol{\Theta}\right)$, we define the loss function of the UL method as 
\begin{equation}
    L(\boldsymbol{\Theta}) = \mathcal{L}_g(\boldsymbol{y}^{L+1}\left(\boldsymbol{\Theta}^{t-1}\right),\boldsymbol{\lambda}),
\end{equation}
which takes into account both the objective function and the constraints of problem \eqref{eq:original}. Then, by using the mini-batch stochastic gradient method, the classic primal-dual algorithm for solving problem \eqref{eq:problem2.1l} can be  written as follows \cite{d4}
\begin{align}\label{eq:unsupervised}
    \boldsymbol{\Theta}^t =\,\,\,& \boldsymbol{\Theta}^{t-1} - \eta^t \Bigg(\frac{1}{B}\sum_{u \in \mathcal {B}}\nabla_{\boldsymbol{\Theta}}f_0\left(\boldsymbol{y}^{L+1}\left(\boldsymbol{\Theta}^{t-1}\right); \boldsymbol{a}^u\right) \\
    & + \sum_{n\in \mathcal{N}} \lambda_n\left(\frac{1}{B}\sum_{u \in \mathcal {B}}\nabla_{\boldsymbol{\Theta}}f_n\left(\boldsymbol{y}^{L+1}\left(\boldsymbol{\Theta}^{t-1}\right); \boldsymbol{a}^u\right)\right)\Bigg) \nonumber \\
    {\lambda}^t_n =\,\,\,& {\lambda}^{t-1}_n + \eta^t \Bigg(\frac{1}{B}\sum_{u \in \mathcal {B}}f_n\left(\boldsymbol{y}^{L+1}\left(\boldsymbol{\Theta}^{t-1}\right); \boldsymbol{a}^u\right)\Bigg)^{+},
\end{align}
where $t$ is the iteration index of the iterative procedure, while $(\cdot)^+$ denotes the operation $\mathrm{max}\{0,\cdot\}$. First, by  updating the training parameters $\boldsymbol{\Theta}$ through back-propagation, the DNN learns to minimize the Lagrangian function of (6). Afterwards, by updating $\boldsymbol{\lambda}$ the dual function given in (7) is maximized. The iterative updates of (9)-(10) provide approximate solutions to the optimal value of $\boldsymbol{x}^{*}$, and to the optimal value of $\boldsymbol{\lambda}^{*}$.

\section{Variable-structure Optimization Problems}
\subsection{Multi-task Learning for Optimization Problems}
\subsubsection{The Multi-task Optimization Mapping}
Despite the potential of SL and UL in addressing optimization problems, the mapping of \eqref{mapping} is influenced by the set $\mathcal{N}$, which reflects the dimensionality of the considered optimization problem and the underlying objective of the optimization task. 
As a consequence, we need to consider a new mapping $\mathcal{F}^{\mathrm{MT}}$ that can simultaneously handle optimization problems of the form of \eqref{eq:original}, but of different dimensionality and objective. As the mapping that represents the solution of an optimization problem of a specific dimensionality or objective can be treated as an individual task, the new mapping should reflect a multi-task operation, involving multiple diverse tasks. Then, the DNN which will be used to approximate the mapping $\mathcal{F}^{\mathrm{MT}}$ should be able to adjust its number of nodes of its input and output layers based on the  dimensionality values of all considered tasks, while balancing the training to generalize its performance to all tasks' objectives. With a slight abuse of notation, we define the set $\mathcal{N}^{\mathrm{MT}} =\{\mathcal{N}_1, \mathcal{N}_2, ..., \mathcal{N}_K\}$, with cardinality $K$, to be the set which contains all sets that reflect the dimensionality values of all $K$ considered tasks.  Moreover, for each $\mathcal{N}_i,\forall i\in \{1,2,...,K\}$, it holds that $\mathcal{N}_i= \{1,2,...,N_i\}, N_i \in \mathbb{Z}^{+}$, where $N_i$ is the input and the output dimensionality of the $i$-th task.
Then, the corresponding mapping of \eqref{mapping} in the multi-task case is given as follows
\begin{equation} \label{mapping2}
\mathcal{F}^{\mathrm{MT}}: \,\, \mathcal{A}_i \xrightarrow[]{\text{\eqref{eq:original}}} \mathcal{X}_i^{*},\,\, \forall \mathcal{N}_i \in \mathcal{N}^{\mathrm{MT}},
\end{equation}
where $\boldsymbol{a}_i\in\mathcal{A}_i$ is the input parameter of the $i$-th task, $\mathcal{A}_i\subseteq\mathbb{R}^{N_i}$, $\mathcal{X}_i\subseteq\mathbb{R}^{N_i}$, $\boldsymbol{x}_i^{*}(\boldsymbol{a}_i)$ is the optimal solution of \eqref{eq:original} for the $i$-th task, and $\mathcal{X}_i^{*}$ is the set containing the optimal values for the $i$-th task, $\forall \boldsymbol{a}_i\in\mathcal{A}_i$, and is defined similar to \eqref{eq:opt_set}. In essence, to create the mapping $\mathcal{F}^{\mathrm{MT}}$, multiple mappings $\mathcal{F}_i$, $\forall \mathcal{N}_i \in \mathcal{N}^{\mathrm{MT}}$, as given in \eqref{mapping}, need to be handled simultaneously by a single DNN. 
\subsubsection{Modular Sharing for MTL}
\begin{figure}[!t] 
\vspace{-0cm}
\centering
\includegraphics[width=1\columnwidth]{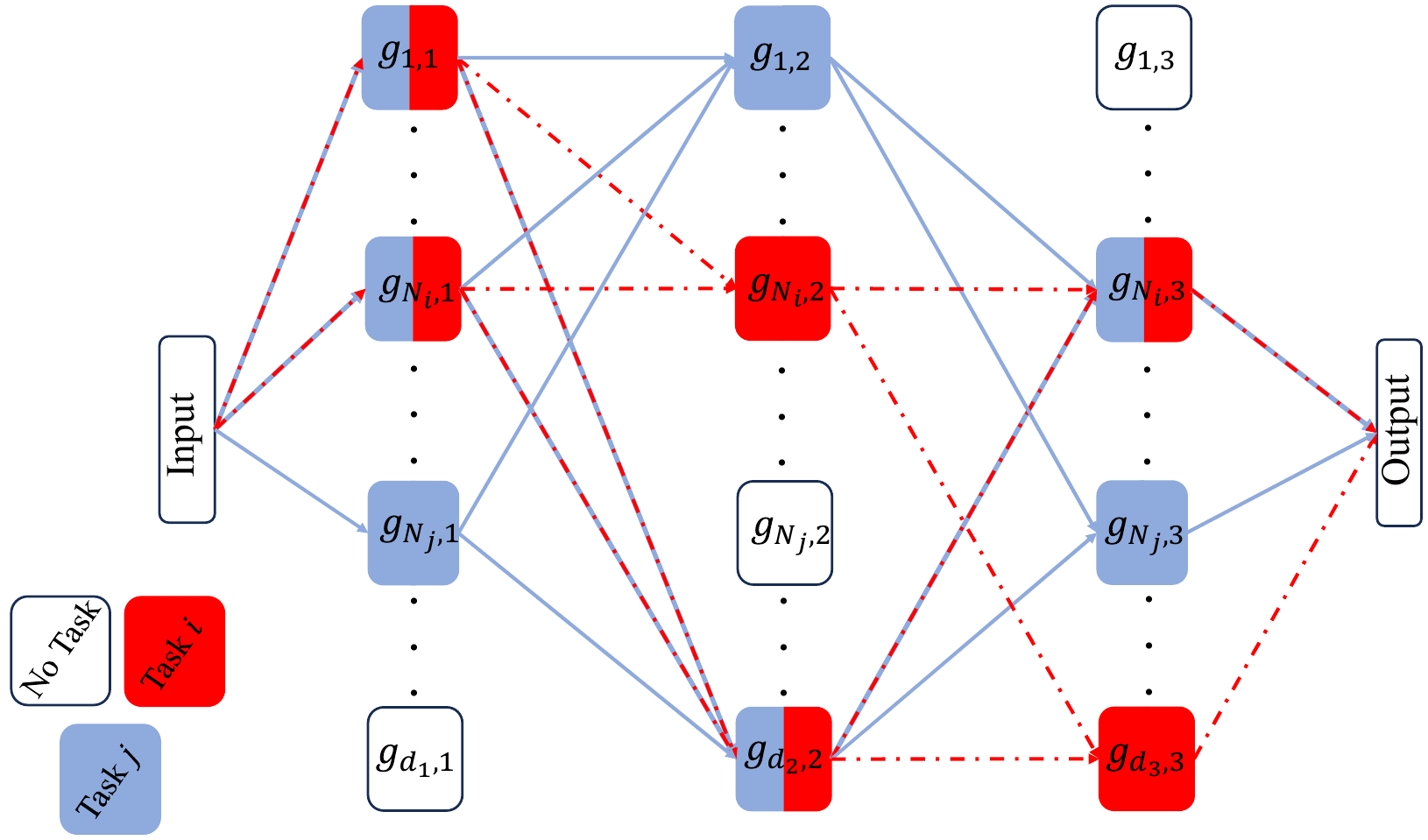}
\centering
\caption{A modular sharing example for MTL.} 
\label{modular_sharing}
\vspace{-0cm}
\end{figure}
Let us consider the computational graph of Fig. \ref{modular_sharing}. We assume that the DNN network has a number of input and output nodes which equal the maximum dimensionality value of all the considered tasks. Then, it has to  hold that
\begin{equation}
    d_1 = d_L = \max\{N_1,...,N_K\},
\end{equation}
while the values of $d_i, \forall i \in \{2,...,L-1\}$, can be chosen in an arbitrary way. We are interested in designing a single DNN which can be used for completing multiple tasks, though, during forward propagation, different tasks may flow through different sub-networks within the same DNN. Specifically, the aforementioned technique is known as \emph{modular sharing} \cite{mt} and enables a DNN to adapt between different tasks, by allowing the tasks to share some common DNN parameters. Let $\boldsymbol{\Theta}_i, \forall i\in\{1,2,...,K\}$ denote the training parameters of the subnet of the single DNN, dedicated to the $i$-th task. 
From the illustrative example in Fig. \ref{modular_sharing}, it can be observed that different tasks flow through some common pathways in the computational graph of the DNN, and thus, each task causes ``interference" to other tasks, in the sense that common weights should be tuned to exhibit satisfactory performance for all the considered tasks.
As such, from the authors' point of view, there are two main challenges in designing the proposed multi-task DNN. The first is to make the DNN adaptable to its input and output dimension, while appropriately choosing the subnet per task to optimize the amount of interference caused to each task, and the second is to select the multi-task loss function to regulate the trade-off of jointly training multiple tasks. 

\begin{figure*}[!t]
\centering
\includegraphics[width=.85\textwidth]{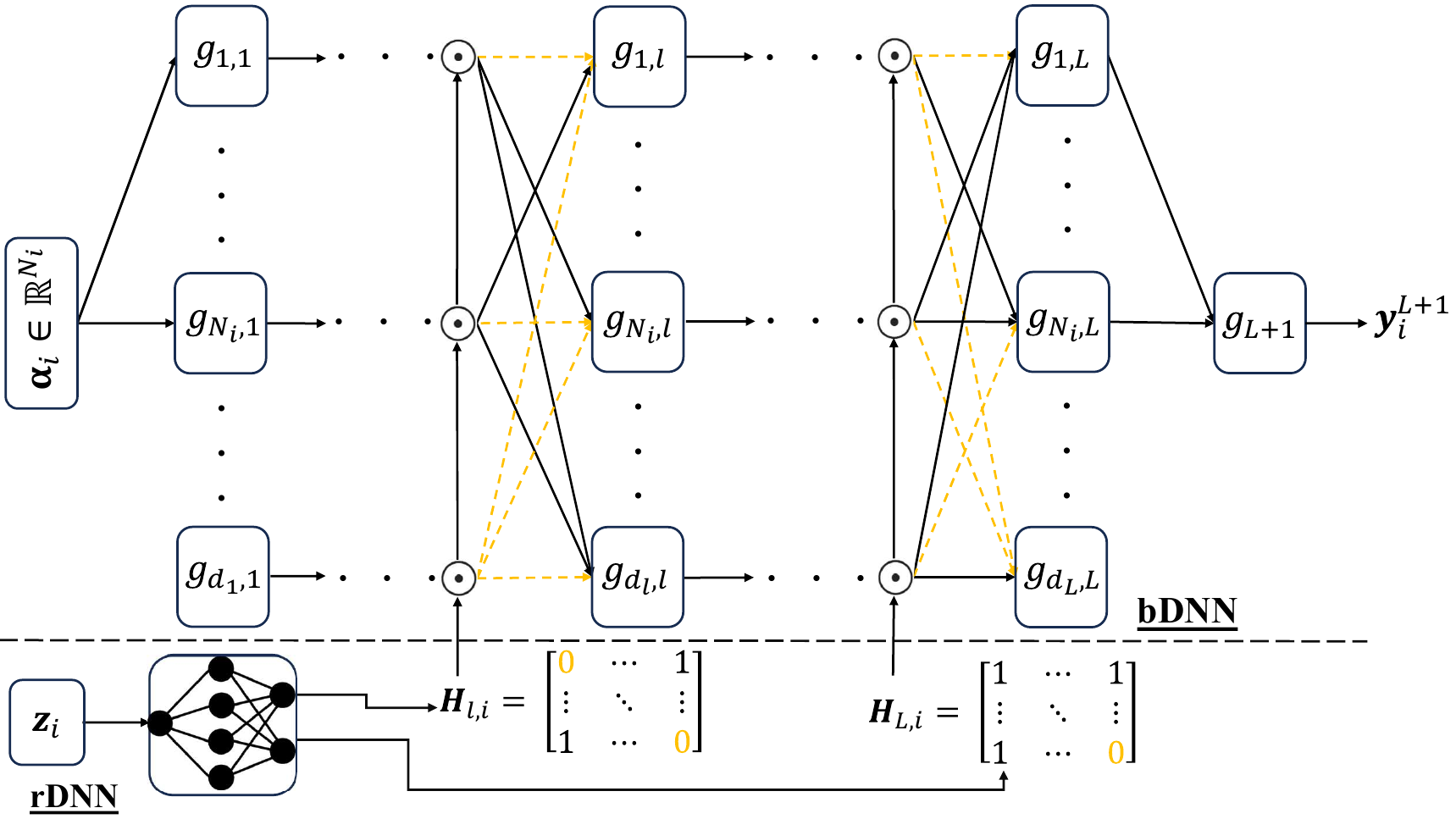}
\centering
\vspace{0cm}
\caption{The proposed multi-task DNN architecture.} 
\label{nn_architecture}
\end{figure*}

\subsection{Conditional Computation with Routing}
Regarding the first challenge, the proposed multi-task DNN can be designed based on \emph{conditional computation} with \emph{routing} \cite{mt,routing}. Conditional computation is a method in which specific paths of the DNN's computational graph are selected during the forward propagation, depending on the input of the network. Specifically, we consider two main DNN components, the bDNN, whose role is to let the input data per task propagate within itself during forward propagation, and the rDNN, which is responsible for dynamically selecting which parts of the bDNN architecture will be activated during the forward propagation of each task \cite{mt,routing}. \textcolor{black}{However, differently from \cite{mt}, we will consider a hard parameter sharing approach which does not increase the number of trainable parameters, thus providing a fair comparison between our scheme and any optimal single-task DNN.} 
For each different task a different DNN, which is a subnet of the bDNN, is instantiated based on the rDNN's decision. Essentially, the rDNN associates tasks with subnets from the bDNN, i.e., with the parameter matrices $\boldsymbol{\Theta}_i$.  As a consequence, the conditional computation-based multi-task DNN is dynamic between inputs and outputs, as well as between tasks, since multiple paths of these dynamically instantiated DNN architectures are shared. 

To make this clearer, let us consider Fig. \ref{nn_architecture}. Each task is assigned to a specific subnet, while the subnet selection depends on the rDNN, which takes as input the parameter $\boldsymbol{z}_i \in \{0,1\}^K$, which is a one-hot vector that indicates the task's index. We denote the output of the rDNN for the $i$-th task as 
\begin{equation}
    \boldsymbol{y}_{\mathrm{r},i}(\boldsymbol{\Phi},\boldsymbol{z}_i) = \{\boldsymbol{H}_{l,i}\}_{l=2}^{L},\,\forall i \in \{1,2,...,K\},
\end{equation}
where $\boldsymbol{\Phi}$ denotes the training parameters of the rDNN. All the outputs of the rDNN, $\{\boldsymbol{H}_{l,i}\}_{l=2}^{L},\,\forall i \in \{1,2,...,K\}$, are binary matrices, with dimension of $d_{l-1}\times d_l$. The rDNN's output, as shown in Fig. 2, is multiplied with the weights of the bDNN element-wisely per hidden layer, providing a differentiable way to design the architecture of each subnet per task. Therefore, the training parameters of the hidden layers, $l=2,...,L$, of the $i$-th subnet which are dedicated to the $i$-th task, are given as
    $\left\{\boldsymbol{H}_{l,i}\odot \boldsymbol{W}_{l},\boldsymbol{b}_{l}\right\}_{l=2}^{L},\,\forall i \in \{1,2,...,K\}$,
where $\odot$ denotes the element-wise product. Due to the binary nature of the matrices $\boldsymbol{H}_{l,i}$, the proposed parameter sharing is called \emph{hard parameter sharing}. 
Therefore, the overall training parameters of the $i$-th subnet, for the $i$-th task, during its forward propagation through the bDNN, are then given as 
\begin{equation}
\begin{aligned}
    \boldsymbol{\Theta}_i =&\Big\{\boldsymbol{W}_1[1:N_i,:],\boldsymbol{b}_1,\left\{\boldsymbol{H}_{l,i}\odot \boldsymbol{W}_{l},\boldsymbol{b}_{l}\right\}_{l=2}^{L},\\
    &\boldsymbol{W}_{L+1}[1:N_i,:],\boldsymbol{b}_{L}\Big\},\,\forall i \in \{1,2,...,K\}.
\end{aligned}
\end{equation}
Essentially, the functionality of the rDNN resembles that of the dropout method, except for the rDNN learning which weights to set to zero, while dropout selects this randomly. Moreover, we note that following \cite{slimmable}, the input and output layer of the bDNN for the $i$-th task do not depend to the rDNN's output, but instead for the $i$-th task, the first $N_i$ neurons of the first layer and of the last layer of the bDNN are selected. This is necessary to ensure that both the input and the output layers of the $i$-th subnet have a number of input and output nodes equal to the dimensionality value of the $i$-th task. This is illustrated in both Fig. \ref{modular_sharing}, and Fig. \ref{nn_architecture}. 
\subsubsection{The rDNN's Activation Function \&  Weight Initialization}
\begin{figure}[t]
\centering
\includegraphics[width=1\columnwidth]{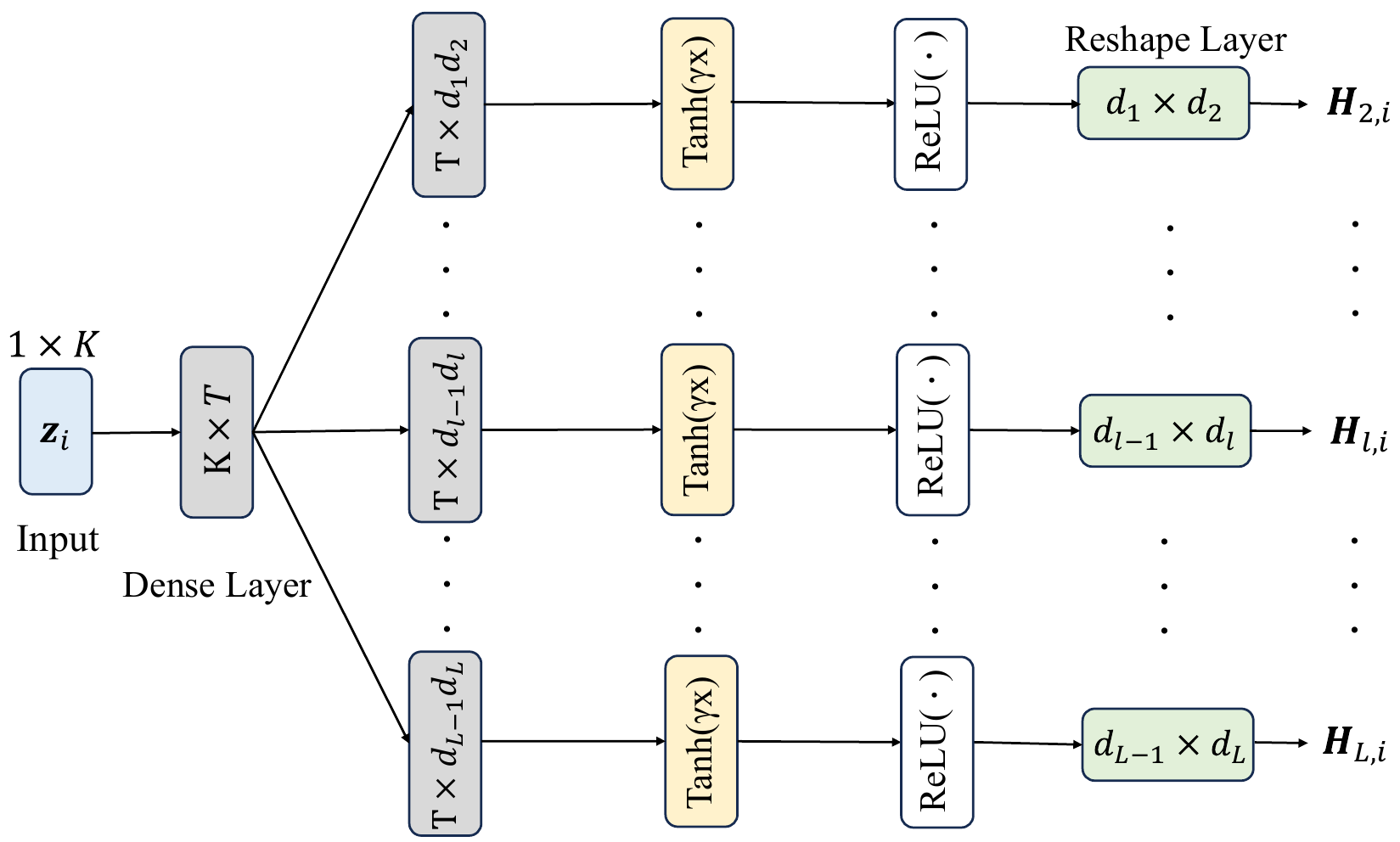}
\centering
\caption{The proposed rDNN architecture.} 
\label{fig:router}
\end{figure}

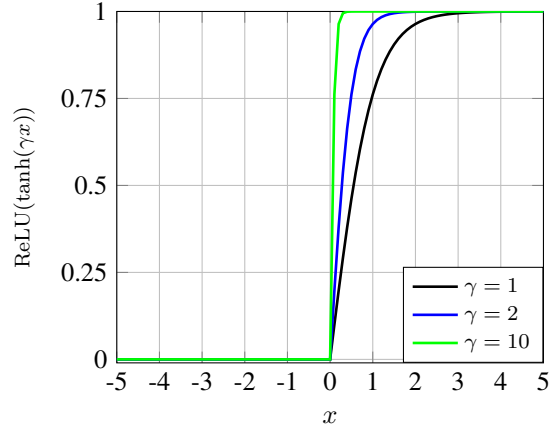
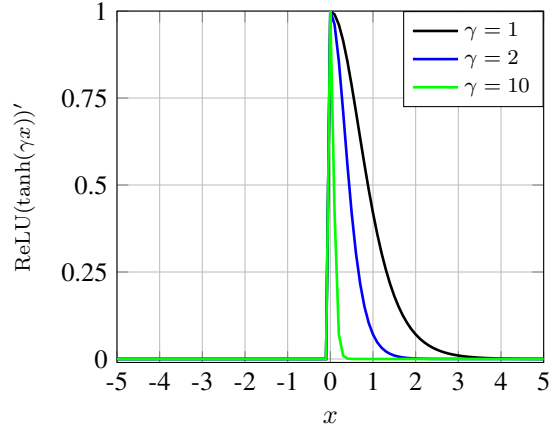
\begin{figure}[t!]
\begin{subfigure}{0.4\textwidth}
    \centering
    \begin{tikzpicture}
        \begin{axis}[
            scaled ticks=false, tick label style={/pgf/number format/fixed},
            width=1\textwidth,
            xlabel = {$x$},
            ylabel = {\footnotesize $\mathrm{ReLU}(\mathrm{tanh}(\gamma x))$},
            ymin = -0.01,
            ymax = 1,
            xmin = -5,
            xmax = 5,
            xtick = {-5,-4,-3,-2,-1,0,1,2,3,4,5},
            xticklabels={-5,-4,-3,-2,-1,0,1,2,3,4,5},
            ytick = {0,0.25,0.5,0.75,1},
            yticklabels = {0,0.25,0.5,0.75,1},
            grid = major,
            legend entries = {{$\gamma=1$},{$\gamma=2$},{$\gamma=10$}},
            legend cell align = {left},
            legend style={at={(1,0)},anchor=south east, font=\footnotesize},
            ]
            \addplot[
            black,
            no marks,
            mark size = 2,
            line width = 1pt,
            style = solid,
            ]
            table {figs/tanh/1_tanh.dat};
            \addplot[
            blue,
            no marks,
            mark size = 2,
            line width = 1pt,
            style = solid,
            ]
            table {figs/tanh/2_tanh.dat};
            \addplot[
            green,
            no marks,
            mark size = 2,
            line width = 1pt,
            style = solid,
            ]
            table {figs/tanh/10_tanh.dat};
        \end{axis}
    \end{tikzpicture}
    \caption{The $\mathrm{ReLU}(\mathrm{tanh}(\gamma x))$ function.}
    \label{fig:tanh}
\end{subfigure}
\begin{subfigure}{0.4\textwidth}
    \centering
    \begin{tikzpicture}
        \begin{axis}[
            scaled ticks=false, tick label style={/pgf/number format/fixed},
            width=1\textwidth,
            xlabel = {$x$},
            ylabel = {\footnotesize $\mathrm{ReLU}(\mathrm{tanh}(\gamma x))'$},
            ymin = -0.01,
            ymax = 1,
            xmin = -5,
            xmax = 5,
            xtick = {-5,-4,-3,-2,-1,0,1,2,3,4,5},
            xticklabels={-5,-4,-3,-2,-1,0,1,2,3,4,5},
            ytick = {0,0.25,0.5,0.75,1},
            yticklabels = {0,0.25,0.5,0.75,1},
            grid = major,
            legend entries = {{$\gamma=1$},{$\gamma=2$},{$\gamma=10$}},
            legend cell align = {left},
            legend style={at={(1,1)},anchor=north east, font=\footnotesize},
            ];
            \addplot[
            black,
            no marks,
            mark size = 2,
            line width = 1pt,
            style = solid,
            ]
            table {figs/tanh_der/1_der.dat};
            \addplot[
            blue,
            no marks,
            mark size = 2,
            line width = 1pt,
            style = solid,
            ]
            table {figs/tanh_der/2_der.dat};
            \addplot[
            green,
            no marks,
            mark size = 2,
            line width = 1pt,
            style = solid,
            ]
            table {figs/tanh_der/10_der.dat};
        \end{axis}
    \end{tikzpicture}
    \caption{The derivative of $\mathrm{ReLU}(\mathrm{tanh}(\gamma x))$}
    \end{subfigure}
    \caption{The $\mathrm{ReLU}(\mathrm{tanh}(\gamma x))$ and its derivative with respect to $\gamma$.}
    \label{fig:tanhh}
    \vspace{-0.5cm}
\end{figure}

The complete architecture of the rDNN is given in Fig. \ref{fig:router}, where $T$ is the number of nodes of the hidden layer, and its value is chosen in Table 2.
We notice that the $\mathrm{tanh}(\gamma x)$ function, for great enough values of $\gamma$, can approximate the step function, while also keeping a non-zero gradient. Therefore, we define the activation function of the last layer to be given as $\mathrm{ReLU}\left(\mathrm{tanh}\left(\gamma x\right)\right)$, which approximately belongs to $\{0,1\}$ for great values of $\gamma$. The proposed activation function and its gradient are shown in Fig. \ref{fig:tanhh}. The activation function has zero gradient at point zero, while its gradient also goes towards zero as the function's input increases. Therefore, if the initial input of the rDNN is non-positive, its output will be zero, while the output will never leave zero, due to the gradient being zero as well. Also, if the rDNN's weight initialization result in a rDNN output value close to two, where the gradient of the chosen activation function is near zero, the rDNN will not be trained, and the rDNN's output will always be 1. As such, the initialization of all the rDNN's weights follows the $ \mathcal{N}\left(\mu\boldsymbol{1}, \sigma^2\boldsymbol{1} \right)$ which is the normal distribution with mean value $\mu$ and variance $\sigma^2$, with $\mu>0$ and $\sigma^2<\mu$, while all the rDNN's biases are set to zero.
The initialization needs to ensure that the initial output of the rDNN will be near one everywhere, while also ensuring that the gradient is non-zero, and ideally near its maximum. Good values of $\mu$ and $\sigma^2$ can be easily found offline by running a few simulations on the rDNN. Essentially, the rDNN is trained to cut the destructive common branches between different tasks, while keeping these which do help the bDNN to converge. However, due to the trade-off that occurs between accurately approximating the step function with the $\mathrm{tanh}(\gamma x)$ function, and its vanishing gradient, the actual output of the rDNN lies in $[0,1]$, with most of its values concentrated near zero or near one. To address this, once the rDNN is jointly trained with the bDNN, the final hard parameter sharing of the rDNN occurs as follows
\begin{equation}
    \boldsymbol{y}_{\mathrm{r},i} = \mathrm{Sign}\left(\mathrm{ReLU}\left(\boldsymbol{y}_{\mathrm{r},i}(\boldsymbol{\Phi},\boldsymbol{z}_i)-0.5\right)\right),\,\forall i \in \{1,2,...K\},
\end{equation}
which forces all outputs of the rDNN which is below 0.5 to go to zero, and all values above 0.5 to go to one. Due to the element-wise product of the rDNN's outputs and the bDNN's weights, the bDNN's weights which are not useful during the forward propagation of a specific task are not utilized, but the same weights can be used during the forward propagation of another task. Similarly, some weights can be used during the forward propagation of multiple tasks, or during the forward propagation of none task. 
\vspace{-0cm}
\subsubsection{The training procedure of the multi-task DNN}
Following the analysis of the previous subsections we define the loss function of the multi-task DNN, including both the SL-based and the UL-based tasks, as follows
\begingroup
\allowdisplaybreaks
\begin{align}
     L(\boldsymbol{\Omega}) & = \underbrace{ \sum_{i\in\mathcal{K}_S}\beta_i \left[\frac{1}{B_{i}}\sum_{u \in \mathcal {B}_i} \mathcal{L}\left(\boldsymbol{x}_i^{*}(\boldsymbol{a}_i^u)-\boldsymbol{y}_i^{L+1}\left(\boldsymbol{\Theta}_i, \boldsymbol{\Phi};\boldsymbol{a}_i^u\right)\right)\right]}_\text{SL-based tasks} \nonumber \\
    & + \underbrace{\sum_{i\in\mathcal{K}_U}\beta_i \left[\frac{1}{B_{i}}\sum_{u \in \mathcal {B}_i}f_{0,i}\left(\boldsymbol{y}_i^{L+1}\left(\boldsymbol{\Theta}_i,\boldsymbol{\Phi}\right); \boldsymbol{a}_i^u\right) \right.}_\text{objective function of UL-based tasks} \nonumber\\
    & \underbrace{\left.+\sum_{n=1}^{N_i} {\lambda}_{n,i}^{t-1}\left(\frac{1}{B_{i}}\sum_{u \in \mathcal {B}_i}f_{n,i}\left(\boldsymbol{y}_i^{L+1}\left(\boldsymbol{\Theta}_i,\boldsymbol{\Phi}\right); \boldsymbol{a}_i^u\right)\right)\right]}_\text{{constraints of UL-based tasks}},    
\end{align}
\endgroup
\textcolor{black}{where $\sum_{i\in\mathcal{K}_U}\beta_i + \sum_{i\in\mathcal{K}_S}\beta_i= 1$, $|\mathcal{K}_U|+|\mathcal{K}_S| = K$, $\mathcal{K}_U$ is the set containing all UL-based tasks, and $\mathcal{K}_S$ is the set containing all SL-based tasks. Moreover, $\boldsymbol{\Omega}\triangleq\{\boldsymbol{\Theta}, \boldsymbol{\Phi}\}$ and it contains all the bDNN's and the rDNN's trainable parameters, $\boldsymbol{a}_i^u$ is the $u$-th parameter sample of the $i$-th task from the dataset $\mathcal{D}_i=\{\boldsymbol{a}_i^u,\boldsymbol{x}_i^{*}(\boldsymbol{a}_i^u)\}_{u=1}^{D_i}$, if $i\in\mathcal{K}_S$ and from the dataset $\mathcal{D}_i=\{\boldsymbol{a}_i^u\}_{u=1}^{D_i}$, if $i\in\mathcal{K}_U$, which consists of  $D_i=\lvert\mathcal{D}_i\rvert$ samples.} Moreover, $\mathcal{B}_i\subseteq\mathcal{D}_i$, $B_i=\lvert\mathcal{B}_i\rvert$ is the batch size used per task, and $\boldsymbol{y}_i^{L+1}$ is the output of the bDNN for the $i$-th task. We note that a major challenge for the UL-based tasks is to provide an output which satisfies all the constraints, of all different tasks. To this end,  a new dual variable, ${\lambda}_{n,i}$, was defined for the $n$-th constraint of the $i$-th UL-based task, while $f_{0,i}$ is the cost function of the $i$-th task, and $f_{n,i}$ is the $n$-th constraint of the $i$-th task. Then, the bDNN and the rDNN are jointly trained until convergence. \textcolor{black}{The gradient step of the primal-dual optimization of the proposed multi-task DNN at round $t$, is given as follows
\begingroup
\allowdisplaybreaks
\begin{align}
        \boldsymbol{\Omega}^t &=\, \boldsymbol{\Omega}^{t-1} -\eta^t \sum_{i\in\mathcal{K}_S}\beta_i \nonumber \\
        &\times\left[\frac{1}{B_i}\sum_{u \in \mathcal {B}_i} \nabla_{\boldsymbol{\Omega}}\mathcal{L}\left(\boldsymbol{x}_i^{*}(\boldsymbol{a}^i_u),\boldsymbol{y}_i^{L+1}\left(\boldsymbol{\Theta}_i,\boldsymbol{\Phi};\boldsymbol{a}_i^u\right)\right)\right] \nonumber \\
        & - \eta^t\sum_{i\in\mathcal{K}_U}\beta_i \nonumber \left[\frac{1}{B_{i}}\sum_{u \in \mathcal {B}_i}\nabla_{\boldsymbol{\Omega}}f_{0,i}\left(\boldsymbol{y}_i^{L+1}\left(\boldsymbol{\Theta}_i,\boldsymbol{\Phi}\right); \boldsymbol{a}_i^u\right) \right.\nonumber \\
        &\left. +\sum_{n=1}^{N_i} {\lambda}_{n,i}^{t-1}\left(\frac{1}{B_{i}}\sum_{u \in \mathcal {B}_i}\nabla_{\boldsymbol{\Omega}}f_{n,i}\left(\boldsymbol{y}_i^{L+1}\left(\boldsymbol{\Theta}_i,\boldsymbol{\Phi}\right); \boldsymbol{a}_i^u\right)\right)\right],\\
        {\lambda}^t_{n,i} &=\, {\lambda}^{t-1}_{n,i} + \eta^t \Bigg(\frac{1}{B_{i}}\sum_{u \in \mathcal {B}_i}f_{n,i}\left(\boldsymbol{y}_i^{L+1}\left(\boldsymbol{\Theta}_i,\boldsymbol{\Phi}\right); \boldsymbol{a}_i^u\right)\Bigg)^{+}.
\end{align}
\endgroup}
It is easy to conclude that when a training parameter $w_{n,l}(k)$ is zero or it is not included to the $i$-th subnet's parameter matrix $\boldsymbol{\Theta}_i$, due to the rDNN, its gradient is also zero with respect to the $i$-th task. Therefore, this parameter does not affect the $i$-th task during the backprogragation of the multi-task bDNN. 
After convergence, the rDNN's outputs $\{\boldsymbol{H}_{l,i}\}_{l=1}^{L-1},\,\forall i \in \{1,2,...,K\}$ are saved and the rDNN is halted from training. Afterwards, the bDNN is retrained. \textcolor{black}{Therefore, the gradient step at round $t$ is given by (17), (18), by updating only with respect to $\boldsymbol{\Theta}$, as follows
\begingroup
\allowdisplaybreaks
\begin{align}
    \boldsymbol{\Theta}^t &= \boldsymbol{\Theta}^{t-1} - \eta^t \sum_{i=1}^{K} \beta_i \nonumber\\
    &\times\left[\frac{1}{B_i}\sum_{u \in \mathcal{B}_i} \nabla_{\boldsymbol{\Theta}} \mathcal{L} \left(\boldsymbol{x}_i^{*}(\boldsymbol{a}^i_u), \boldsymbol{y}_i^{L+1} \left(\boldsymbol{\Theta}_i; \boldsymbol{a}_i^u\right) \right)\right] \nonumber \\
    &- \eta^t \sum_{i \in \mathcal{K}_U} \beta_i \left[\frac{1}{B_i} \sum_{u \in \mathcal{B}_i} \nabla_{\boldsymbol{\Omega}} f_{0,i} \left( \boldsymbol{y}_i^{L+1} \left( \boldsymbol{\Theta}_i \right); \boldsymbol{a}_i^u \right) \right. \nonumber \\
    & + \left. \sum_{n=1}^{N_i} {\lambda}_{n,i}^{t-1} \left( \frac{1}{B_i} \sum_{u \in \mathcal{B}_i} \nabla_{\boldsymbol{\Omega}} f_{n,i} \left( \boldsymbol{y}_i^{L+1} \left( \boldsymbol{\Theta}_i \right); \boldsymbol{a}_i^u \right) \right) \right].
\end{align}
\endgroup
\begin{equation}
            {\lambda}^t_{n,i} =\, {\lambda}^{t-1}_{n,i} + \eta^t \left(\frac{1}{B_{i}}\sum_{u \in \mathcal {B}_i}f_{n,i}\left(\boldsymbol{y}_i^{L+1}\left(\boldsymbol{\Theta}_i\right); \boldsymbol{a}_i^u\right)\right)^{+}.
\end{equation}}

\begin{algorithm}[t]
\linespread{0.9}\selectfont
\begin{algorithmic}[1]\label{alg1}
\caption{Training of the proposed multi-task DNN} 
\State {$\boldsymbol{\Phi}_\mathrm{w}\sim \mathcal{N}\left(\mu\boldsymbol{1}, \sigma^2\boldsymbol{1} \right)$,
\,\,$\boldsymbol{\Phi_\mathrm{b}} = 0$}
\State{$\boldsymbol{\Theta \sim \text{Xavier distribution}}$, $\boldsymbol{\Omega}= \left\{\boldsymbol{\Theta},\boldsymbol{\Phi}\right\}$}
\State{\textbf{Initialize} $\mathcal{D}_i, \beta_i, \forall i \in \mathcal{K}_U,\mathcal{K}_S$, $t_1,\,t_2$, $\eta$, $\mathrm{scheme}$}
\While {$t \leq t_1$} 

    \State{Update $\boldsymbol{\Omega}^t$ according to (17)}
    \State{Update $\lambda_{n,i}^t,\,\forall n,i$, according to (18)}
    \State {$t\gets t+1$}
\EndWhile
\State{Output $ \boldsymbol{y}_{\mathrm{r},i}(\boldsymbol{\Phi},\boldsymbol{z}_i),\,\forall i \in \mathcal{K}_U,\mathcal{K}_S$}
\State{$\boldsymbol{y}_{\mathrm{r},i}\!\gets\!\mathrm{Sign}\left(\mathrm{ReLU}\left(\boldsymbol{y}_{\mathrm{r},i}(\boldsymbol{\Phi},\boldsymbol{z}_i)-0.5\right)\right),\,\forall i \in \mathcal{K}_U,\mathcal{K}_S$}
\State{$\boldsymbol{\Phi}$ is halted from training}
\State{$\boldsymbol{\Theta \sim \text{Xavier distribution}}$}
\State{$t \gets 0$}
\While {$t\leq t_2$} 
    \State{Update only $\boldsymbol{\Theta}^t$ according to (19)}
    \State{Update $\lambda_{n,i}^t,\,\forall n,i$, according to (20)}
    \State {$t\gets t+1$}
\EndWhile
\State{Output $\boldsymbol{\Theta}^*$, $\lambda_{n,i},\,\forall n,i$, $\boldsymbol{y}_{\mathrm{r},i},\,\forall i \in \mathcal{K}_U,\mathcal{K}_S$}
\end{algorithmic}
\end{algorithm}
The complete training procedure of the   UL-based multi-task DNN is also given in Algorithm 1.
In Algorithm 1, we begin by initializing all weights randomly. Specifically, the bDNN's weights are initiated using the Xavier distribution. The primary aim of Xavier initialization is to set the weights so that the variance of weights which belong to different layers is equal across the DNN. This uniform variance plays a crucial role in preventing the issues of gradient explosion or vanishing. Subsequently, both the bDNN and the rDNN undergo joint training until the multi-task loss function converges. The outcome of this training process is the desired rDNN, which associates a subnet of the bDNN with a specific task. Following this phase, the rDNN is halted from training.
Then, the bDNN is retrained until convergence, since retraining the bDNN was observed to slightly improve the performance of the proposed multi-task scheme, compared to the case when the bDNN is only jointly trained with the rDNN. We note that when dealing with a small number of tasks, it is feasible to train on all tasks simultaneously at each step, using the SGD per task. However, this approach may not be practical when the number of tasks increases significantly. In such cases, it is preferable to randomly sample a subset of tasks to train the DNN on per round \cite{mt}, while a uniform task selection can maintain equal bias across all subnets selection \cite{once}. The final output of Algorithm 1 consists of both the routing information and the bDNN's parameters.

\subsection{Complexity Analysis}
In this section we present the complexity of the proposed scheme, while a comparison with the widely used interior-point method (IPM) is given. For the IPM the complexity is given as $\mathcal{O}(\log(1/\epsilon)(N+m)^{3})$, since approximately $\log(1/\epsilon)$ number of iterations are needed for the IPM method to reach a feasible solution of accuracy $\epsilon$, while each iteration requires a Newton step which involves the solution of the modified system of KKT equations . We note that $m$ is the number of constraints, where in our case $m = N$, while it has also been shown that $\log(1/\epsilon)$ scales according to $\sqrt{m}$, therefore the IPM's complexity for a convex optimization problem of \eqref{eq:original} is of the order of $\mathcal{O}(N^{3.5})$ \cite{boyd2004convex}.

For a feedforward NN (FNN) the complexity is given as the sum of the forward and the backward pass. The forward-pass complexity of a single DNN is dominated by the weight matrix multiplication cost, as it can be seen in (3). Thus the forward-pass complexity of one data sample is given as $\mathcal{O}(\sum_{l=1}^Ld_l\times d_{l+1})$. Given that the backward-pass has the same computational cost with the forward-pass \cite{marco}, the overall back propagation algorithm for training a DNN is of the order of $\mathcal{O}(I_\mathrm{E}D\sum_{l=1}^Ld_l\times d_{l+1})$, where $I_\mathrm{E}$ is the number of training epochs, and $D$ is the datasize as defined previously. For simplicity and following (12), we can assume that $d_l = \max\{N_1,...,N_K\} = N,\, \forall l \in \{1,...,L\}$. Then the forward-pass complexity of one sample is given as $\mathcal{O}(LN^2)$ which is quadratic with respect to (w.r.t.) the dimensionality of the tasks, and it is linear w.r.t. the number of hidden layers. We note then, that using an appropriate number of hidden layers, a DNN network with inference complexity lower than that of the IPM can be achieved, while using techniques like vectorization and GPU acceleration can further decrease the inference time of the DNN.


Next, we have to generalize this analysis to the proposed multi-task DNN. First, we note that following the previous analysis, to train $K$ distinct DNNs for $K$ different tasks requires a complexity of $\mathcal{O}(KI_\mathrm{E}DLN^2)$. 
To find the complexity of the proposed MTL scheme, we make the assumption that data from all $K$ tasks are fed into the DNN during each feed-forward pass; nonetheless, this is not always the case. The complexity of the rDNN during a feed-forward pass can be given according to $\mathcal{O}(K^2TLN^2)$, but the input to the rDNN is a one-hot vector, thus containing multiplications with 0s which do not need to be taken into account, resulting in a complexity of $\mathcal{O}(KTLN^2)$. 
After the rDNN's feed-forward pass, a Hadamard product takes place, as shown in Fig. 2. This product is of complexity $\mathcal{O}(LN^2)$. The complexity of the bDNN is the same as that of an FNN. Therefore, the total complexity of the proposed scheme during training equals $
\mathcal{O}(I_\mathrm{E}D(LN^2 + KTLN^2 + LN^2)) = \mathcal{O}(I_\mathrm{E}DLKTN^2)$,
which differs only by a constant $T$ from that of training $K$ distinct DNNs, showcasing the efficiency in terms of complexity of the proposed multi-task scheme.
We note that the inference complexity of this scheme is considerably smaller in practice, since not all $K$ tasks need to be inferred at the same time, while the rDNN's output, following (13), will contain many zeros, which considerably reduces the complexity of the Hadamard product and the inference of the bDNN per task.

\section{Application to Wireless Resource Management}
In this section, we consider two wireless resource management applications where the effectiveness of the proposed MTL mechanism will be evaluated. First, we present a delay minimization problem that serves as the use case for evaluating the performance of the multi-task SL method. Second, we formulate the average sum capacity maximization problem while accounting for average power constraints, which will be employed to assess the effectiveness of the multi-task UL method. 
\textcolor{black}{Then, the multi-task DNN will be jointly trained to both problems across various dimensionality values, following (16). Although both involve power allocation, the two problems exhibit fundamentally different optimal solutions. For average sum capacity maximization, a water-filling like approach should be employed which allocates more power to the best channels to maximize throughput. In contrast, delay minimization needs a different strategy, as allocating more power to the best channels worsens overall delay by increasing the delay for users or subchannels with poor conditions. This contrast highlights the conflicting nature of the two tasks and it will be used to demonstrate the multi-task capability of the proposed DNN, which will have to learn two distinct and opposing power allocation strategies across diverse network configurations.}

\subsection{FDMA Delay Minimization via SL}
FDMA finds numerous applications in wireless networks \cite{d7,d13}, and it is anticipated that 6G networks will leverage the FDMA technology \cite{6gkaragian}. Additionally, both Wi-Fi 6 and Wi-Fi 7, will depend on \emph{channel bonding} \cite{wifi6,wifi7}. Channel bonding represents a specific implementation paradigm of FDMA, enabling the aggregation of multiple frequency channels to enhance bandwidth and data rates within the Wi-Fi network. Therefore, the minimization of the delay of an FDMA transmission is a pertinent and timely topic, and it will be used to assess the performance of the supervised multi-task scheme. Subsequently, we formulate the problem of minimizing the average delay of a downlink FDMA transmission as follows:
 
\begin{equation}\tag{$\textbf{P}_3$}\label{eq:delay_min}
    \begin{aligned}
    \underset{\boldsymbol{P}}{\mathrm{min}}&\quad
    \frac{1}{N}\sum_{n=1}^{N} \frac{L_n}{ W\log_2\left(1+\frac{P_{n}|h_n|^2}{N_0W} \right)} \\
    \,\,\textbf{s.t.} \,\,\, \mathrm{C}_1:& \quad\sum_{n=1}^{N} P_{n} \leq P_{\mathrm{tot}},\\
     \mathrm{C}_2:& \quad 0 \leq P_n \leq P_{\mathrm{tot}},\,\forall n \in \{1,...,N\},
    \end{aligned}
\end{equation}
where $N$ is the number of available frequency bands, $P_n$ is the transmission power of the $n$-th frequency band, $P_\mathrm{tot}$ is the maximum transmission power of the transmission, $|h_n|^2$ is the channel gain, $L_n$ is the data size to be transmitted to the $n$-th band, $N_0$ is additive white Gaussian noise (AWGN), and $W$ is the available bandwidth. Problem \eqref{eq:delay_min} is convex and can be solved optimally, thus the SL approach can be utilized. Therefore, given the feature vector $\boldsymbol{h} = [h_1,...,h_N]$, the respected optimal power allocation $\boldsymbol{P}^{*} = [P_1^{*},...,P_N^{*}]$ can be found. Thus, by solving problem \eqref{eq:delay_min} $D$ times, the dataset $\mathcal{D} = \{\boldsymbol{h}^u,\boldsymbol{P}^{*,u}\}_{u=1}^{D}$ of size $D$ can be created. This dataset will be created for all $\mathcal{N}_i$, $\forall i \in \{1,...,K\}$, under investigation. The bDNN architecture follows that of Fig. 2, where all hidden layers are dense, followed by the rectified linear unit (ReLU) function, with exception to the last layer, which is followed by the softmax activation function, such that constraints $\mathrm{C}_1$ and $\mathrm{C}_2$ of problem \eqref{eq:delay_min} are satisfied.  

\begin{table}[t!]
\linespread{0.75}\selectfont
\centering
\caption{Simulation Parameters}
\begin{tabular}{|P{1.15cm}||P{2.75cm}||P{2.75cm}|}
\hline
\textbf{Parameter} & \textbf{FDMA delay minimization} & \textbf{Average sum capacity maximization}\\ 
\hline
$P_\mathrm{tot}$& $10$ W &   $1$ W  \\  
\hline
$P_\mathrm{av}$& -- &   $0.5$ W  \\  
\hline
$W$&  $1$ MHz &  --  \\  
\hline
$N_0$& $-174$ dBm/Hz &   $-174$ dBm/Hz  \\ 
\hline
$L_n$& $1$ Mbit &   --  \\ 
\hline\hline
\multicolumn{2}{|c||}{\textbf{DNN's parameters}} & \textbf{Value} \\ \hline
\multicolumn{2}{|c||}{$K$} & 7  \\  
\hline
\multicolumn{2}{|c||}{$N_i$} & $\{5,8,10,12,15,18,20\}$ \\  
\hline
\multicolumn{2}{|c||}{$D$} & $40\,\mathrm{k}$   \\  
\hline
\multicolumn{2}{|c||}{$L+1$} & 7  \\  
\hline
\multicolumn{2}{|c||}{$T$} & 20  \\  
\hline
\multicolumn{2}{|c||}{$d_1,d_6$} & 20  \\  
\hline
\multicolumn{2}{|c||}{$d_2 - d_5$} & $32$  \\  
\hline
\multicolumn{2}{|c||}{$\gamma$} & $5$    \\  
\hline
\multicolumn{2}{|c||}{$\mu$} & $0.1$    \\  
\hline
\multicolumn{2}{|c||}{$\sigma$} & $0.001$   \\  
\hline
\multicolumn{2}{|c||}{$B_i$} & $32$   \\  
\hline
\multicolumn{2}{|c||}{$\eta$} & $0.001$    \\  
\hline
\multicolumn{2}{|c||}{$t_1$} & $5$ k  \\  
\hline
\multicolumn{2}{|c||}{$t_2$} & $5$ k  \\  
\hline
\multicolumn{2}{|c||}{$\beta_i$} & $1/K$   \\  
\hline
\end{tabular}
\label{tab:example}
\end{table}

\subsection{Average Sum Capacity Maximization via UL}
Following \cite{d4} we will formulate the average sum capacity maximization problem, under an average and max power constraint in order to verify the proposed UL-based multi-task approach. The problem is given below
\begin{equation}\tag{$\textbf{P}_4$}\label{eq:sum_capacity}
    \begin{aligned}
    \underset{\boldsymbol{P}}{\mathrm{max}}&\quad 
    \mathbb{E}_{\boldsymbol{h}}\left[ \log\left(1+\sum_{n=1}^N|h_i|^2P_i\right)\right] \\
    \,\,\textbf{s.t.}\,\,\, \mathrm{C}_1:&\quad \mathbb{E}_{\boldsymbol{h}}\left[\sum_{n=1}^NP_i\right]\leq P_\mathrm{av}\\
    \mathrm{C}_2:& \quad 0 \leq P_n \leq P_{\mathrm{tot}},\,\forall n \in \{1,...,N\}.
    \end{aligned}
\end{equation}
We observe that \eqref{eq:sum_capacity} is a stochastic optimization problem, making it notably more complex to address in the general scenario. Nevertheless, the expectation within the objective function and the constraints can be readily managed using the SDG method as presented in (9) \cite{d4}. Hence, there is no requirement for further analysis, and problem (P4) can be resolved using either (9) in the single-task case, or alternatively, (22)-(23) in the multi-task scenario. Consequently, the proposed multi-task approach remains applicable to stochastic optimization problems as well. In contrast to the SL approach, the dataset comprises only of the feature vectors $\boldsymbol{h}$, thus, $\mathcal{D} = \{\boldsymbol{h}^u\}_{u=1}^{D}$. This dataset will be also established for all, $\mathcal{N}_i$, $\forall i \in \{1,...,K\}$. In this case, the bDNN architecture also follows that of Fig. 2, where all hidden layers are dense, followed by the ReLU function, with exception of the last layer, which is followed by the Sigmoid activation function, such that the constraint $\mathrm{C}_2$ of problem \eqref{eq:sum_capacity} always hold. The constraint $\mathrm{C}_1$ will be forced to hold via the primal-dual optimization.

\begin{table*} [h]
    \linespread{0.8}\selectfont
    \centering
    \caption{Evaluation of the SL case, \scalebox{0.9}{$\mathbb{E}_{\boldsymbol{h}} \left[f_0\left(\left(\boldsymbol{y}^{L+1}\left(\boldsymbol{\Theta};\boldsymbol{h}\right)\right)-f_0\left(\boldsymbol{x}^{*};\boldsymbol{h}\right)\right)^2\right]$}}
    \label{table:Table1}
    \begin{tabular}{|c|c|c|c|c|c|c|c|}
    \hline
    \diagbox[width=12em]{\textbf{Scheme}}{\textbf{Task\vspace{-0.3cm}}}& $\mathbf{N=5}$ & $\mathbf{N=8}$ & $\mathbf{N=10}$ & $\mathbf{N=12}$ & $\mathbf{N=15}$ & $\mathbf{N=18}$ & $\mathbf{N=20}$ \\
    \hline
\textbf{single-task DNN} & 0.0404 & 0.0243 & 0.0218 & 0.0203 & 0.0193 & 0.0176 & 0.0152 \\
\hline
\textbf{proposed multi-task DNN} & 0.0425 & 0.0351 & 0.0353 & 0.0287 & 0.0247 & 0.0203 & 0.0171 \\
\hline
\textbf{naive multi-task DNN} & 0.2405 & 0.1865 & 0.1380 & 0.0965 & 0.0689 & 0.0510 & 0.0428 \\
\hline
\textbf{zero-padding} & 0.2216 & 0.1025 & 0.0659 & 0.0434 & 0.0264 & 0.0183 & 0.0178 \\
\hline
\end{tabular}
\end{table*}
\begin{table*} [t]
\linespread{0.8}\selectfont
    \centering
    \caption{Evaluation of the UL case, average sum capacity (bps/Hz)}
    \label{table:Table1}
    \begin{tabular}{|c|c|c|c|c|c|c|c|}
    \hline
    \diagbox[width=12em]{\textbf{Scheme}}{\textbf{Task\vspace{-0.3cm}}}& $\mathbf{N=5}$ & $\mathbf{N=8}$ & $\mathbf{N=10}$ & $\mathbf{N=12}$ & $\mathbf{N=15}$ & $\mathbf{N=18}$ & $\mathbf{N=20}$ \\
    \hline
    \textbf{single-task DNN} & 2.236 & 2.797 & 3.1053 & 3.263 & 3.515 & 3.754 & 3.883 \\
    \hline
    \textbf{proposed multi-task DNN} & 2.233 & 2.783 & 3.047 & 3.244 & 3.5086 & 3.7309 & 3.863 \\
    \hline
    \textbf{naive multi-task DNN} & 2.075 & 2.5064 & 2.7618 & 2.9531 & 3.2135 & 3.4437 & 3.5729 \\
    \hline
    \textbf{zero-padding} & 0.9166 & 1.4712 & 1.8420 & 2.2043 & 2.76 & 3.3107 & 3.6823 \\
    \hline
    \end{tabular} 
\end{table*}
\begin{figure}[t!]
\begin{subfigure}{0.5\textwidth}
    \centering
    \begin{tikzpicture}
        \begin{axis}[
            scaled ticks=false, tick label style={/pgf/number format/fixed},
            width=0.8\linewidth,
             xlabel = {Iteration index $t$},
            ylabel = {\scalebox{.75}{$\mathbb{E}_{\boldsymbol{h}} \left[\left(f_0\left(\boldsymbol{y}^{L+1}\left(\boldsymbol{\Theta}^t;\boldsymbol{h}\right)\right)-f_0\left(\boldsymbol{x}^{*};\boldsymbol{h}\right)\right)^2\right]$}},
            ymin = 0.01,
            ymax = 1,
            xmin = 0,
            xmax = 5000,
            xtick={0,1000,2000,3000,4000,5000},
            xticklabels={0,1000,2000,3000,4000,5000},
            ytick = {0,0.2,...,1},
            grid = major,
            legend entries = {{single-task DNN},{zero-padding},{naive multi-task DNN}, {proposed multi-task DNN}},
            legend cell align = {left},
            legend style={at={(1,1)},anchor=north east, font=\scriptsize},
            ]
            \addplot[
            lightgray,
            no marks,
            mark size = 2,
            line width = 1pt,
            style = solid,
            ]
            table {figs/N_5_supervised/test_opt_2.dat};
            \addplot[
            blue,
            no marks,
            mark size = 2,
            line width = 1pt,
            style = solid,
            ]
            table {figs/N_5_supervised/zp_5_2.dat};
            \addplot[
            green,
            no marks,
            mark size = 2,
            line width = 1pt,
            style = solid,
            ]
            table {figs/N_5_supervised/naive_5_new_2.dat};
            \addplot[
            yellow,
            no marks,
            mark size = 2,
            line width = 1pt,
            style = solid,
            ]
            table {figs/N_5_supervised/mod_shar_2.dat};
        \end{axis}
    \end{tikzpicture}
    \caption{Evaluation of $N=5$.}
    \label{fig:6}
\end{subfigure}
\begin{subfigure}{0.5\textwidth}
    \centering
    \begin{tikzpicture}
        \begin{axis}[
            scaled ticks=false, tick label style={/pgf/number format/fixed},
            width=0.8\linewidth,
            xlabel = {Iteration index $t$},
            ylabel = {\scalebox{.75}{$\mathbb{E}_{\boldsymbol{h}} \left[\left(f_0\left(\boldsymbol{y}^{L+1}\left(\boldsymbol{\Theta}^t;\boldsymbol{h}\right)\right)-f_0\left(\boldsymbol{x}^{*};\boldsymbol{h}\right)\right)^2\right]$}},
            ymin = 0.02,
            ymax = .5,
            xmin = 0,
            xmax = 5000,
            xtick={0,1000,2000,3000,4000,5000},
            xticklabels={0,1000,2000,3000,4000,5000},
            ytick = {0,0.1,...,.5},
            yticklabels = {0,0.1, 0.2,0.3,0.4,0.5},
            grid = major,
            legend entries = {{single-task DNN},{zero-padding},{naive multi-task DNN}, {proposed multi-task DNN}},
            legend cell align = {left},
            legend style={at={(1,1)},anchor=north east, font=\scriptsize},
            ]
            \addplot[
            lightgray,
            no marks,
            mark size = 2,
            line width = 1pt,
            style = solid,
            ]
            table {figs/N_12_supervised/test_opt_2.dat};
            \addplot[
            blue,
            no marks,
            mark size = 2,
            line width = 1pt,
            style = solid,
            ]
            table {figs/N_12_supervised/zp_12_2.dat};
            \addplot[ 
            green,
            no marks,
            mark size = 2,
            line width = 1pt,
            style = solid,
            ]
            table {figs/N_12_supervised/naive_12_new_2.dat};
            \addplot[
            yellow,
            no marks,
            mark size = 2,
            line width = 1pt,
            style = solid,
            ]
            table {figs/N_12_supervised/mod_shar_2.dat};
        \end{axis}
    \end{tikzpicture}
    \caption{Evaluation of $N=12$.}
    \label{fig:6}
\end{subfigure}
\hfill
\begin{subfigure}{0.5\textwidth}
    \centering
    \begin{tikzpicture}
        \begin{axis}[
            scaled ticks=false, tick label style={/pgf/number format/fixed},
            width=0.8\linewidth,
            xlabel = {Iteration index $t$},
            ylabel = {\scalebox{.75}{$\mathbb{E}_{\boldsymbol{h}} \left[\left(f_0\left(\boldsymbol{y}^{L+1}\left(\boldsymbol{\Theta}^t;\boldsymbol{h}\right)\right)-f_0\left(\boldsymbol{x}^{*};\boldsymbol{h}\right)\right)^2\right]$}},
            ymin = 0,
            ymax = .15,
            xmin = 0,
            xmax = 5000,
            xtick={0,1000,2000,3000,4000,5000},
            xticklabels={0,1000,2000,3000,4000,5000},
            ytick = {0,0.05, 0.10, 0.15},
            yticklabels = {0,0.05, 0.10, 0.15},
            grid = major,
            legend entries = {{single-task DNN},{zero-padding},{naive multi-task DNN}, {proposed multi-task DNN}},
            legend cell align = {left},
            legend style={at={(1,1)},anchor=north east, font=\scriptsize},
            ]
            \addplot[
            lightgray,
            no marks,
            mark size = 2,
            line width = 1pt,
            style = solid,
            ]
            table {figs/N_20_supervised/test_opt_2.dat};
            \addplot[
            blue,
            no marks,
            mark size = 2,
            line width = 1pt,
            style = solid,
            ]
            table {figs/N_20_supervised/zp_20_2.dat};
            \addplot[
            green,
            no marks,
            mark size = 2,
            line width = 1pt,
            style = solid,
            ]
            table {figs/N_20_supervised/naive_20_new_2.dat};
            \addplot[
            yellow,
            no marks,
            mark size = 2,
            line width = 1pt,
            style = solid,
            ]
            table {figs/N_20_supervised/mod_shar_2.dat};
        \end{axis}
    \end{tikzpicture}
    \caption{Evaluation of $N=20$.}
    \label{fig:6}
\end{subfigure}
\caption{The SL-based multi-task performance.}
\end{figure}
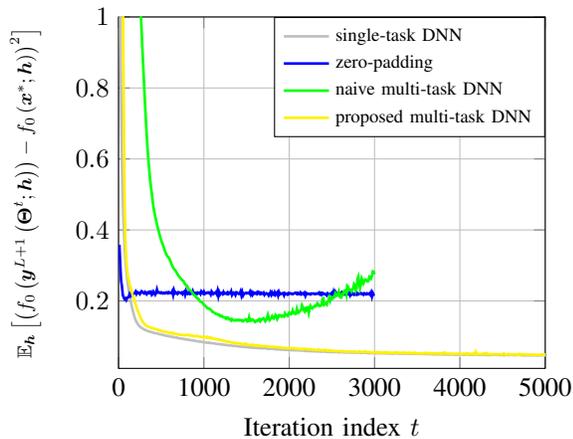
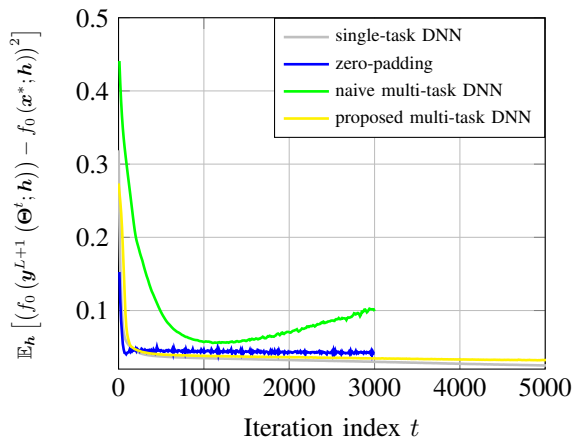
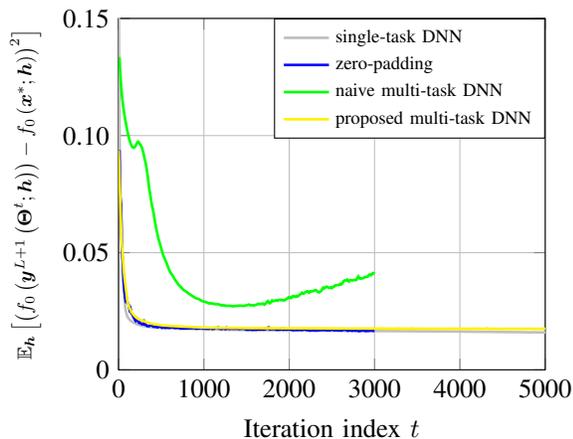
\section{Numerical Simulations}
In this section, the proposed multi-task approach is evaluated. All parameters are given in Table 2. The results were averaged using a Monte Carlo approach over 100 iterations, while both resource allocation scenarios were studied under Rayleigh fading. The Adam optimizer was used to train all the illustrated schemes, following Algorithm 1. The training dataset consists of $30\,k$ samples, while the testing dataset consists of $10\,k$ samples. Both datasets were created based on Rayleigh fading and section IV. \textcolor{black}{We note that two problems, with different objective and constraints, for seven different setups, i.e., dimensionality values, were chosen. It is note, that both problems have essentially different solutions when their dimensionality changes, since the structure of the optimization problem changes. Then, problems of different dimensionality can be considered as different tasks, which was explained in section III}. Thus, the total number of individual tasks amounts to fourteen. \textcolor{black}{All considered multi-task schemes were jointly trained to all tasks, but for clarity, the evaluation of individual tasks will be shown per graph. The evaluation occurred by using the testing datasets of each individual task during various training iteration indexes.} Moreover, the proposed multi-task scheme was designed containing the smallest number of parameters which achieve a satisfactory testing evaluation. Subsequently, we ensured that all benchmarks mentioned below also utilize the same number of parameters: 
\begin{itemize}
    \item \emph{single-task DNN \cite{d4}}: The single-task DNN has an overall number of training parameters equal to the number of the bDNN's training parameters. A different single-task DNN is trained for each different task. \textcolor{black}{Moreover, the single-task DNN of \cite{d4} has been proven theoretically optimal for unsupervised optimization problems, thus its performance can be considered as the upper (lower) bound for the performance of the proposed multi-task DNN.}
    \item \emph{zero-padding}: ZP has been used in the literature to address the issue of dynamic input and output of DNNs, especially in the case of wireless resource management  \cite{z1,z10,z12}. The architecture and the number of training parameters of the ZP-based DNN is the same with the single-task DNN, but the dataset of each task is padded with zeros until the feature and label dimensions of all tasks become equal to  $\max\{N_1,...,N_K\}$. Moreover, the dataset is such that it contains a different real value, i.e., an index value, which differs between tasks. This way, a single DNN can learn to separate different tasks and to be trained subject to all tasks. Thus, this ZP scheme is a multi-task benchmark too. We note that ZP may not always be applicable to UL-aided optimization problems, but it is applicable to problem \eqref{eq:sum_capacity}. This is in contrast to the proposed multi-task approach which is always applicable to all UL-aided problems. 
    \item \emph{naive multi-task DNN \cite{mitsiou2024multi}}: In contrast to the proposed multi-task scheme, the router is not trained, but its output is one everywhere. Thus, all tasks cause interference to each other, since all tasks utilize the same nodes of all hidden layers of the bDNN. \textcolor{black}{Nonetheless, the router still enables the bDNN to dynamically adjust its input and output layer according the task's dimensionality. Thus, the weights of the naive scheme are given as \scalebox{0.9}{$\boldsymbol{\Theta}_i =\Big\{\boldsymbol{W}_1[1:N_i+1,:],\boldsymbol{b}_1,\left\{\boldsymbol{W}_{l},\boldsymbol{b}_{l}\right\}_{l=2}^{L},
    \boldsymbol{W}_{L+1}[1:N_i,:]$,} \scalebox{0.9}{$\boldsymbol{b}_{L}\Big\},\,\forall i \in \{1,2,...,K\}$}. Furthermore, the dataset contains integer indexes per each considered task which are used to separate tasks of equal dimensionality but of different objective, thus it holds that $d_1 = \max\{N_1,...,N_K\}+1$.}
\end{itemize}

In Fig. 5, the performance of the multi-task scheme for SL-based tasks with $N = 5, 12, 20$ is shown. The performance for all $N$ is given in Table 3. The value $\mathbb{E}_{\boldsymbol{h}} \left[f_0\left(\left(\boldsymbol{y}^{L+1}\left(\boldsymbol{\Theta};\boldsymbol{h}\right)\right)-f_0\left(\boldsymbol{x}^{*};\boldsymbol{h}\right)\right)^2\right]$ is plotted. This represents the mean square error of the minimum delay produced by the DNN-based power allocation, with respect to (w.r.t.) the optimal delay obtained by solving the convex optimization problem in \eqref{eq:delay_min}. First, it can be seen that the single-task DNN has the best performance, which verifies the claim that the comparison between the multi-task DNN and all single-task DNNs is fair. It can be seen that the ZP cannot generalize its performance in the multi-task case, since for $N=5$, and other $N$ values from Table 3, it converges poorly compared to the single-task DNN. Moreover, it is observed that its learning stops earlier than the other schemes. \textcolor{black}{On the other hand, the naive multi-task scheme is shown to perform even worse, due to the increased interference between all tasks. Its testing performance resembles overfitting, nonetheless, the training curve was also observed to have a similar behavior, thus it is concluded that due to the interference between all tasks the naive scheme cannot generalize its performance.} We note that since ZP employs multiple zeros to its input layer, the interference between the tasks is expected to be lower which is verified from its better performance compared to the naive scheme. \textcolor{black}{Interestingly, the proposed multi-task approach performs almost equally to the single-task DNN, despite the interference between tasks. This is due to the fact that the rDNN cuts down any unnecessary interaction between different tasks, enabling the bDNN to generalize its performance to all tasks.} Furthermore, we note that the dataset for the task with $N = 5$ was observed to have the biggest variance compared to all considered tasks' datasets, which justifies the fact that $N=5$ has the greatest loss function between all tasks, since it is the most challenging task. 

\begin{figure}[h]
    \centering
    \begin{subfigure}[t!]{0.44\textwidth}
        \centering
        \includegraphics[width=\textwidth]{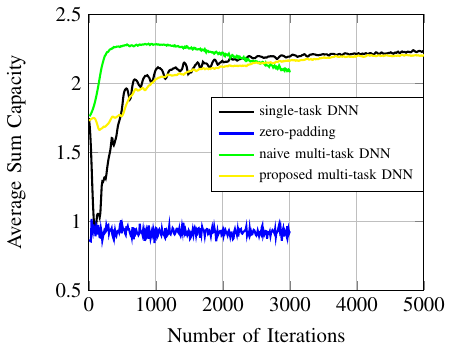}
        \caption{The average sum capacity.}
        \label{fig:subfig1}
    \end{subfigure}
    \hfill
    \begin{subfigure}[t!]{0.44\textwidth}
        \centering
        \includegraphics[width=\textwidth]{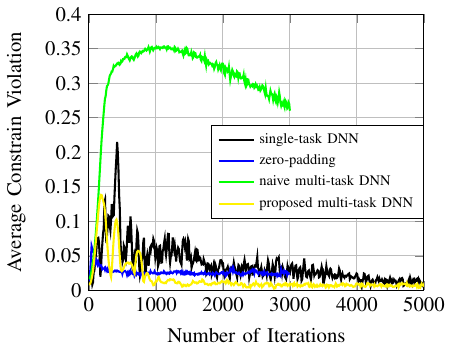}
        \caption{The constraint violation.}
        \label{fig:subfig2}
    \end{subfigure}
    \caption{The UL-based multi-task performance for $N=5$.}
    \label{fig:vert_subfigures}
\end{figure}
In Fig. 6, the performance of the UL-based task, for $N = 5$, is plotted. The figures show the average sum capacity of each task, and the average constraint violation. The single-task DNN outperforms all multi-task DNNs, but its average constraint violation converges slower compared to the proposed multi-task DNN. This can be attributed to the fact that interference between different tasks might accelerate the convergence of the respected single tasks components \cite{once}. Moreover, it is observed that the naive scheme captures more effectively the objective function of the optimization problem, but fails to capture a solution with small constraint violation. In contrary, the ZP scheme gives a solution with small constraint violation, but a low average sum capacity as well. Nonetheless, the proposed multi-task DNN follows the performance of the single-task scheme.

\begin{figure}[t]
    \centering
    \begin{subfigure}[t!]{0.44\textwidth}
        \centering
        \includegraphics[width=\textwidth]{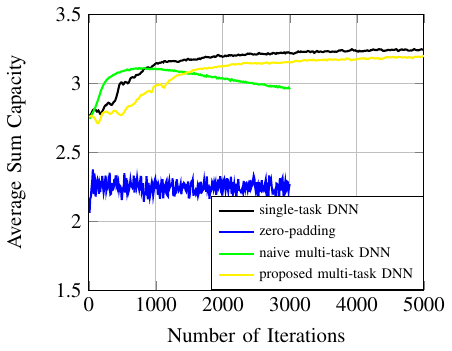}
        \caption{The average sum capacity.}
        \label{fig:subfig1}
    \end{subfigure}
    \hfill
    \begin{subfigure}[t!]{0.44\textwidth}
        \centering
        \includegraphics[width=\textwidth]{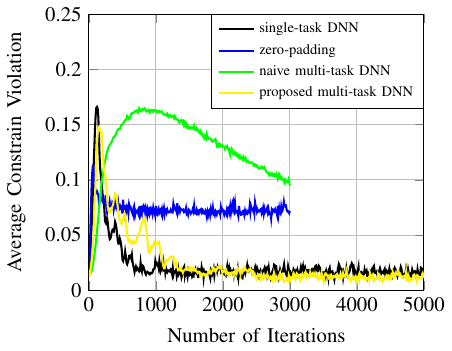}
        \caption{The constraint violation.}
        \label{fig:subfig2}
    \end{subfigure}
    \caption{The UL-based multi-task performance for $N=12$.}
    \label{fig:vert_subfigures}
    \vspace{-.72cm}
\end{figure}
\begin{figure}[t]
    \centering
    \begin{subfigure}[h]{0.44\textwidth}
        \centering
        \includegraphics[width=\textwidth]{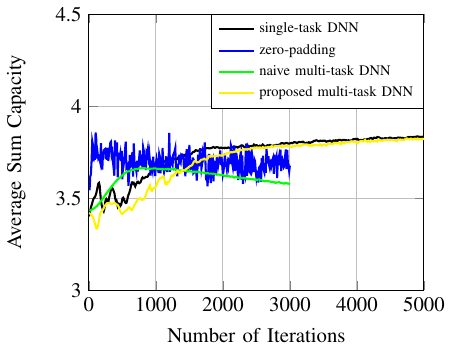}
        \caption{The average sum capacity.}
        \label{fig:subfig1}
    \end{subfigure}
    \hfill
    \begin{subfigure}[h]{0.44\textwidth}
        \centering
        \includegraphics[width=\textwidth]{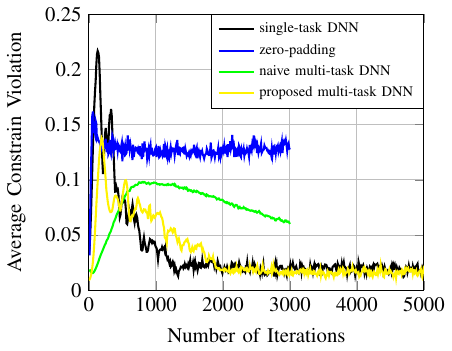}
        \caption{The constraint violation.}
        \label{fig:subfig2}
    \end{subfigure}
    \caption{The UL-based multi-task performance for $N=20$.}
    \label{fig:vert_subfigures}
    \vspace{-.5cm}
\end{figure}
Finally, in Fig. 7 and Fig. 8 the convergence of the testing evaluation for the cases of $N= 12$ and $N=20$ are shown. A similar behavior to the case of $N = 5$ is observed. It is noted that the naive scheme reduces the average constraint violation, but the objective function value reduces as well, thus the naive scheme fails to learn to maximize the objective function. This can be attributed to the fact that the naive scheme cannot mitigate the interference between different tasks. The ZP scheme, for $N = 20$ obtains a value for the average sum capacity which is near to the value of the single-task DNN, but with increased constraint violation. Again, the proposed multi-task scheme has a convergence behavior similar to that of the single-task DNN which showcases the effectiveness of the proposed MTL framework. Furthermore, from the Table 4 it is shown that as the number of users increase the multi-task scheme's obtained value for the average sum capacity also increases, which aligns with the theoretic behavior of average sum capacity maximization in wireless networks.  
\section{Conclusion}
To enable DNNs to handle optimization problems of varying structure and dimensionality, we adopted a MTL approach. By treating optimization problems of different dimensionality, objectives and constraints as distinct tasks, we utilized conditional computation with routing to create a unified DNN architecture capable of jointly addressing all considered optimization problems. The proposed DNN consists of the rDNN, responsible for deciding which nodes and layers of the bDNN are used for each task, while the bDNN contains all potential training parameters directly influencing each task during both forward and backward propagation. Simulation results validated the efficiency of the proposed multi-task scheme. Future directions will aim to integrate the proposed MTL scheme with GNNs for a scalable heterogeneous resource allocation scheme tailored to the complex needs of future wireless networks.
\bibliographystyle{IEEEtran}	
\bibliography{bib.bib}
\end{document}